\documentclass{article} 
\usepackage{colm2024_conference}
\usepackage{scrextend}
\usepackage{booktabs}
\usepackage{graphicx}
\usepackage{enumitem}
\usepackage{wrapfig}
\usepackage{algorithm}
\usepackage{algpseudocode}
\usepackage{natbib}
\usepackage{makecell}
\usepackage{bbm}
\usepackage{array}
\usepackage{amsmath} 
\usepackage{amssymb}
\usepackage{amsfonts}
\usepackage{multirow}
\usepackage{verbatim}
\usepackage{caption}
\usepackage{longtable}
\usepackage{supertabular}
\usepackage{hyperref}
\usepackage{CJKutf8}
\usepackage[utf8]{inputenc} 
\usepackage[T1]{fontenc} 
\usepackage[french,vietnamese,mongolian,greek,english]{babel}
\usepackage{pifont}
\usepackage{afterpage}
\usepackage{tablefootnote}
\usepackage{xspace}
\usepackage{textcomp}
\usepackage{lscape} 
\usepackage{siunitx}
\usepackage{listings}
\usepackage[table]{xcolor} 
\usepackage{adjustbox}

\definecolor{dt}{gray}{0.7}
\definecolor{tongyi-purple}{RGB}{97,92,237}
\colorlet{tongyi-purple-alpha}{tongyi-purple!38}

\usepackage{bbding}
\usepackage{fontawesome}
\usepackage{tgpagella}
\usepackage{latexsym}
\usepackage{microtype}
\definecolor{mydarkblue}{rgb}{0,0.08,0.45}
\definecolor{citecolor}{HTML}{0071BC}
\usepackage{url}            
\usepackage{nicefrac}       
\usepackage{changepage}
\usepackage{xargs}          
\usepackage{lipsum}
\usepackage{subcaption}
\usepackage{endnotes}

\usepackage{pgfplots}
\usetikzlibrary{pgfplots.groupplots}
\pgfplotsset{compat=1.3}
\usepackage{tikz}
\usetikzlibrary{patterns}

\usepackage[most]{tcolorbox}
\usepackage{fvextra}
\usepackage[capitalize,noabbrev]{cleveref}
\crefname{section}{Section}{\S\S}
\Crefname{section}{Section}{\S\S}
\crefname{table}{Table}{Tables}
\crefname{figure}{Figure}{Figures}
\crefname{algorithm}{Algorithm}{}
\crefname{equation}{eq.}{}
\crefname{appendix}{Appendix}{}
\crefformat{section}{Section #2#1#3}
\usepackage{multicol}
\usepackage{fancyvrb} 
\newsavebox{\myverbcontent}
\usepackage{titlesec}
\titleformat*{\section}{\large\bfseries}

\usepackage{nicematrix} 
\usepackage{arydshln}

\makeatletter
\DeclareRobustCommand\onedot{\futurelet\@let@token\@onedot}
\def\@onedot{\ifx\@let@token.\else.\null\fi\xspace}

\title{S1-VL: Scientific Multimodal Reasoning Model with Thinking-with-Images}

\author{
\bf ScienceOne AI}

\begin{document}

\maketitle

\begin{abstract}
We present S1-VL, a multimodal reasoning model for scientific domains that natively supports two complementary reasoning paradigms: \textit{Scientific Reasoning}, which relies on structured chain-of-thought, and \textit{Thinking-with-Images}, which enables the model to actively manipulate images through Python code execution during reasoning. In the Thinking-with-Images mode, the model generates and executes image-processing code in a sandbox environment, obtains intermediate visual results, and continues reasoning in a multi-turn iterative manner. This design is particularly effective for challenging scenarios such as high-resolution scientific chart interpretation, microscopic image understanding, and geometry-assisted reasoning. To construct the training data, we collect scientific multimodal datasets spanning six disciplines: mathematics, physics, chemistry, astronomy, geography, and biology. We further develop a six-dimensional quality filtering framework for reasoning trajectories. To mitigate redundant, ineffective, and erroneous visual operations commonly found in existing datasets, we propose a multi-stage filtering pipeline together with an adaptive data routing strategy. This strategy converts samples with low visual information gain into pure Reasoning-mode data, enabling the model to learn when image operations are truly necessary. S1-VL is trained through a four-stage progressive pipeline: (1) general scientific multimodal supervised fine-tuning (SFT), (2) Thinking-with-Images cold-start SFT combined with curated scientific hard examples, (3) scientific reinforcement learning with SAPO, and (4) Thinking-with-Images reinforcement learning with SAPO. We build S1-VL-32B on top of Qwen3-VL-32B-Thinking and evaluate it on 13 benchmarks. Experimental results show that S1-VL-32B achieves state-of-the-art performance on all five Thinking-with-Images benchmarks, including HRBench-4K, HRBench-8K, MME-RealWorld-CN, MME-RealWorld-Lite, and V*, and outperforms compared systems on scientific reasoning benchmarks such as Physics and VRSBench. Model weights are publicly available on HuggingFace and ModelScope.
\end{abstract}

\begin{figure}[h]
    \centering
    \includegraphics[width=0.9\linewidth]{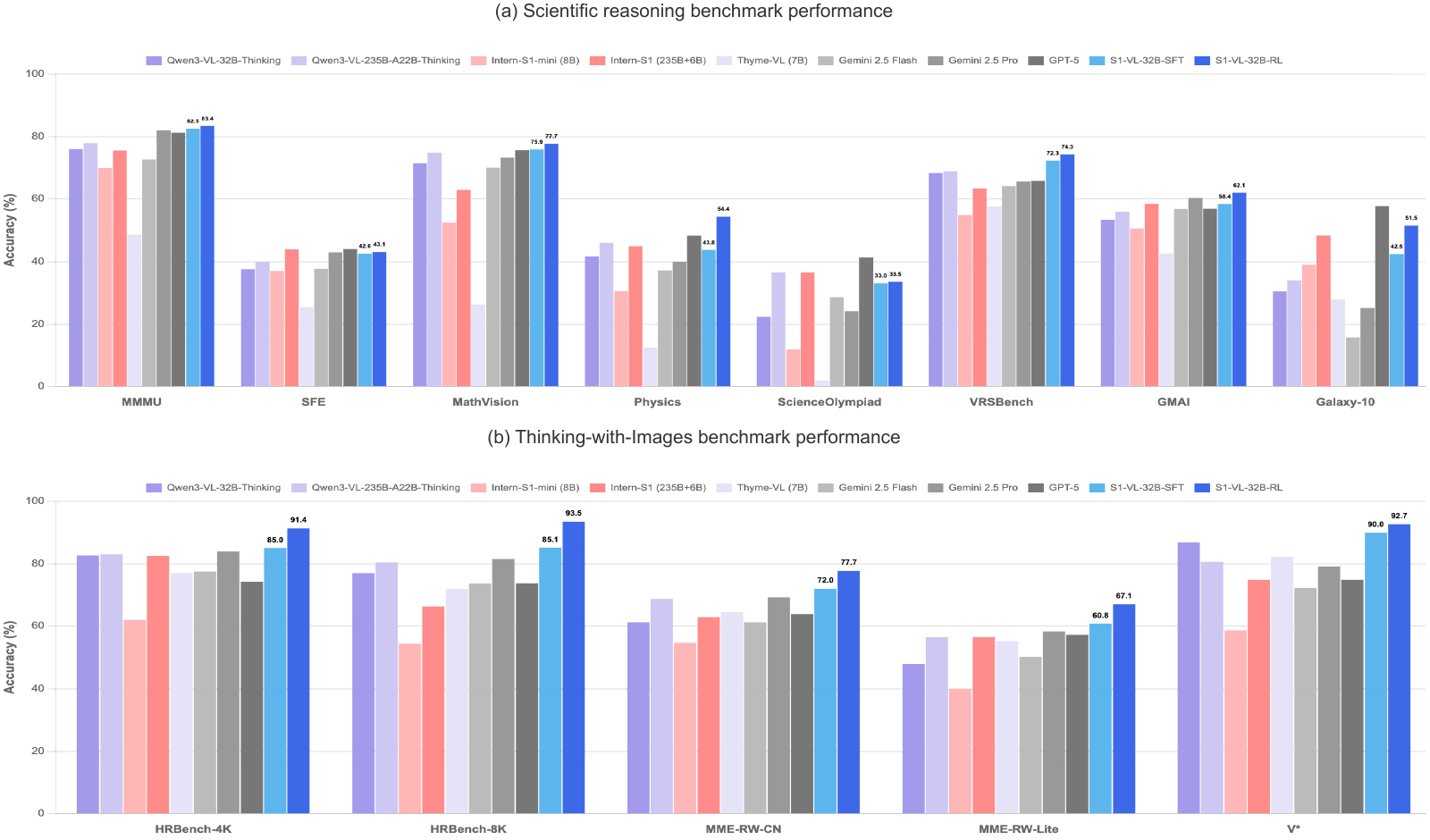}
    \caption{Benchmark performance of S1-VL-32B.}
    \label{fig:eval}
\end{figure}

\newpage
\section{Introduction}
\label{sec:intro}

Multimodal Large Language Models (MLLMs) have achieved remarkable progress in general visual understanding~\citep{team2023gemini, internvl, fu2024vita, bai2025qwen25vltechnicalreport, seed2025seed1_5vl}. Inspired by text-centric reasoning models~\citep{gpt-o1, qwq, guo2025deepseek}, a growing body of work has further explored bringing long-chain reasoning into multimodal settings through supervised fine-tuning~\citep{peng2025skywork, deng2025openvlthinker} or reinforcement learning~\citep{huang2025vision, liu2025VisualARFT, meng2025mm}. However, these models still struggle with fine-grained visual reasoning in scientific domains, where images are characterized by high information density and critical local details. A fundamental limitation is the static treatment of visual input: the image is encoded once as a fixed context and cannot be iteratively inspected or manipulated during reasoning, making these models ill-suited for scientific tasks that demand repeated visual examination and step-by-step analysis.

To overcome the limitations of static visual understanding, the \emph{Thinking-with-Images} paradigm has recently attracted widespread attention~\citep{twi25, su2025thinking}. This paradigm enables models to actively invoke image operations—such as cropping, zooming, and contrast enhancement—during the reasoning process, interleaving visual perception with intermediate reasoning steps and demonstrating strong potential on high-resolution and detail-sensitive tasks. OpenAI's o3~\citep{o3-o4-mini} was among the first to showcase this capability, and a series of open-source efforts have since emerged to explore various instantiations of the paradigm~\citep{zhang2025thyme, zheng2025deepeyes, su2025openthinkimg, qiao2025v, zhang2025skywork}. However, we observe that existing work still suffers from two key limitations. First, in terms of the reasoning paradigm, current methods typically support only a single mode: some models are restricted to standard long-chain textual reasoning without the ability to actively manipulate images~\citep{yang2025qwen3}, while others employ reward mechanisms that indiscriminately encourage tool invocation across all scenarios~\citep{zhang2025thyme, zheng2025deepeyes}, lacking adaptive judgment on whether image operations are truly necessary for a given problem. Second, in terms of training data quality, existing work generally lacks systematic evaluation of the effectiveness of image operations—redundant operations, uninformative intermediate images, formatting errors, and logically inconsistent trajectories are pervasive, substantially undermining training effectiveness and making it difficult for models to genuinely learn when and how to appropriately invoke image operations.

To address these challenges, we propose \textbf{S1-VL}, a powerful multimodal reasoning model tailored to scientific domains that natively supports two complementary reasoning paradigms:

\begin{itemize}[itemindent=0pt, labelsep=4pt, leftmargin=*]
\item \textbf{Scientific Reasoning}: For standard scientific problems across mathematics, physics, chemistry, astronomy, geography, biology, and related domains, the model adopts a structured, multi-step reasoning paradigm. It jointly models and analyzes both textual and visual information, progressively decomposing problems to arrive at solutions. This mode emphasizes end-to-end logical reasoning, enabling robust handling of complex calculations, symbolic derivations, and cross-modal information integration.

\item \textbf{Thinking-with-Images}: For more complex scientific problems that require fine-grained observation and localized analysis, we encourage the model to actively invoke code during the reasoning process. Within a Python sandbox, it performs image operations such as cropping, zooming, contrast enhancement, auxiliary annotations, and keypoint marking to obtain intermediate visual evidence, and iteratively refines its reasoning based on updated inputs. This mode upgrades passive perception into an active ``visual manipulation + reasoning'' loop, significantly enhancing the model’s ability to understand and utilize detailed visual information.
\end{itemize}

To enable this capability, we design a four-stage, progressively structured training pipeline: general scientific multimodal supervised fine-tuning, Thinking-with-Images cold-start fine-tuning with curated scientific hard-example trajectories, SAPO-based scientific reinforcement learning, and SAPO-based Thinking-with-Images reinforcement learning. Together, these stages gradually and systematically develop both strong scientific reasoning and effective image interaction abilities. To support the curriculum-based training strategy, we construct the training data through a unified, multi-stage pipeline aligned with all capabilities. In particular, for Thinking-with-Images, we design a multi-stage quality filtering framework that performs fine-grained trajectory selection across six dimensions, including format compliance, reasoning consistency, image validity, image–text semantic alignment, key information completeness, and cross-turn redundancy. We further introduce an adaptive data routing strategy that converts samples for which image operations provide little meaningful information gain into pure Scientific Reasoning training data. This design enables the model to learn whether image operations are necessary, rather than treating them as an unconditional default.

Built upon Qwen3-VL-32B-Thinking, we train \textbf{S1-VL-32B} and conduct systematic evaluations on 13 multimodal and scientific reasoning benchmarks. As shown in the Figure~\ref{fig:eval}, the model achieves state-of-the-art performance across a comprehensive set of benchmarks. On scientific reasoning tasks, S1-VL-32B demonstrates substantial improvements, with performance approaching that of leading mainstream models. On Thinking-with-Images tasks, S1-VL-32B achieves state-of-the-art results on all five benchmarks, including HRBench-4K, HRBench-8K, MME-RealWorld-CN, MME-RealWorld-Lite, and V*, and outperforms compared systems on scientific reasoning benchmarks such as Physics and VRSBench. These results demonstrate that our approach significantly enhances the model's understanding and fine-grained perception in more complex, detail-oriented scientific scenarios, while further strengthening its interpretability and reasoning capabilities through image manipulation.

\section{Related Work}
\label{sec:related}

\subsection{Multimodal Reasoning with LLMs}
\label{sec:related_mllm}

Multimodal large language models (MLLMs) have progressed from early visual question answering systems to increasingly capable multimodal reasoning models. Early benchmarks in this line of work mainly focused on visual understanding and short-form question answering~\citep{yin2024survey}. More recent datasets have shifted attention toward knowledge-intensive and reasoning-heavy scenarios. For example, ScienceQA~\citep{lu2022learn} integrates scientific knowledge with multimodal question answering, MathVista~\citep{lu2023mathvista} emphasizes mathematically grounded visual reasoning, and OlympiadBench~\citep{he2024olympiadbench} further raises the difficulty to olympiad-level problem solving. These benchmarks collectively reflect a broader trend: multimodal systems are increasingly expected not only to perceive images, but also to reason over them in domain-specific and cognitively demanding settings.

In parallel, model architectures have become substantially more capable of handling complex visual inputs. Representative examples include the InternVL~\citep{zhu2025internvl3} and Qwen-VL~\citep{qwenvl} families, which improve performance on charts, diagrams, and document-like images through stronger visual encoders and tighter vision-language integration. More recently, Qwen3-VL~\citep{bai2025qwen3} combines multimodal perception with chain-of-thought reasoning and demonstrates strong performance on scientific and mathematical benchmarks, making it a strong foundation for scientific multimodal reasoning.

Despite these advances, most existing MLLMs still treat images as static inputs: the image is encoded once, and reasoning proceeds only in the language space thereafter. This design is sufficient for many coarse-grained tasks, but it is fundamentally limited in scientific settings, where successful problem solving often depends on iterative visual inspection, local detail extraction, and the dynamic use of visual evidence. These limitations motivate a more interactive reasoning paradigm in which image understanding is not a one-shot process, but an integral part of the reasoning loop.

\subsection{Thinking-with-Images}
\label{sec:related_twi}

Thinking-with-Images~\citep{twi25, su2025thinking} has recently emerged as a promising paradigm for overcoming the static-image bottleneck in multimodal reasoning. Instead of treating the image as a passive input, this paradigm allows the model to actively manipulate visual content during reasoning through operations such as cropping, zooming, annotation, and auxiliary line drawing. By interleaving visual operations with textual deliberation, the model can iteratively gather task-relevant evidence and refine its reasoning trajectory. This capability is particularly valuable for high-resolution and detail-sensitive tasks, where a single global visual encoding is often insufficient.

Recent work has explored different instantiations of this paradigm. Thyme~\citep{zhang2025thyme} enables image interaction through Python-based tool use and incorporates both execution correctness and answer correctness into reinforcement learning. V-Thinker~\citep{qiao2025v} relies on supervised training over curated multi-step trajectories containing visual operations such as cropping and annotation, allowing the model to gradually acquire tool-use competence. Skywork-R1V4~\citep{zhang2025skywork} further frames visual interaction as an agent-style alternating process between image operations and language reasoning, achieving strong results on high-resolution benchmarks. DeepEyes~\citep{zheng2025deepeyes} departs from the cold-start paradigm and shows that such capabilities can also emerge through end-to-end reinforcement learning alone. From a different angle, Zooming without Zooming~\citep{wei2026zooming} questions the necessity of explicit tool use at inference time and proposes to distill the benefit of region-level inspection into the model during training.

While these studies demonstrate the promise of Thinking-with-Images, the quality of training data remains an underexplored bottleneck. Existing datasets often contain redundant visual operations, uninformative intermediate images, formatting errors, and logically inconsistent trajectories. As a result, models may learn spurious associations between tool use and reasoning, rather than genuinely learning when visual operations are necessary and how they should be integrated into the reasoning process. In this work, we explicitly address this issue through a multi-stage quality filtering pipeline and an adaptive data routing strategy, which together improve both the usefulness and the faithfulness of Thinking-with-Images supervision.

\subsection{Reinforcement Learning for Reasoning Models}
\label{sec:related_rl}

Reinforcement learning has become a central technique for improving reasoning ability in large language models. Recent work such as DeepSeek-R1~\citep{guo2025deepseek} shows that strong reasoning performance can emerge from large-scale outcome-based RL, even without manually annotated reasoning traces. This line of research suggests that, given an appropriate reward signal, models can learn effective reasoning strategies directly from exploration and optimization.

At the algorithm level, GRPO (Group Relative Policy Optimization)~\citep{shao2024deepseekmath} has become a widely adopted method for reasoning-oriented RL because it avoids the need for a separate value model and reduces training complexity. However, its hard-clipping update mechanism can be brittle in practice. When a sequence contains a small number of highly off-policy tokens, clipping may suppress gradient updates for the entire sequence, leading to inefficient learning and instability. This issue is especially problematic in Mixture-of-Experts architectures~\citep{dai2024deepseekmoe}, where token-level routing can further amplify policy variance.

SAPO (Soft Adaptive Policy Optimization)~\citep{sapo} addresses these limitations by replacing hard clipping with a smooth adaptive gating mechanism. Rather than discarding learning signals at the sequence level, SAPO selectively down-weights only highly off-policy tokens, thereby preserving more informative gradients and improving training stability. Prior results show that SAPO achieves stronger optimization behavior and better sample efficiency than GRPO under comparable budgets.

Extending RL from text-only reasoning to multimodal reasoning is nontrivial. In multimodal settings, reward design must account not only for final answer correctness, but also for the validity of tool use, the utility of intermediate visual operations, and the coherence of multi-turn reasoning trajectories. Moreover, when rollouts involve executable code and intermediate image generation, the training pipeline becomes significantly more complex and more vulnerable to reward hacking~\citep{hu2025reward}. In S1-VL, we adopt SAPO in both the scientific reasoning stage and the Thinking-with-Images stage, as it offers a practical balance of stability, efficiency, and robustness for multimodal scientific reasoning.

\section{Model Overview}
\label{sec:model}
\subsection{Base Model}
\label{sec:model_base}
S1-VL is built upon the Qwen3-VL model series without modifying the underlying model architecture. Qwen3-VL provides a strong foundation for scientific multimodal reasoning. It natively supports interleaved image-text contexts of up to 256K tokens and is available in multiple scales ranging from 2B to 235B, covering both dense and mixture-of-experts variants. It also demonstrates strong performance on challenging multimodal benchmarks such as MMMU~\citep{yue2024mmmu}, MathVista~\citep{lu2023mathvista}, and MathVision~\citep{wang2024measuring}. These properties make it particularly suitable for our setting, where the model must handle high-resolution visual inputs, perform multi-step reasoning, and support long-context multimodal interaction.

In this work, we train S1-VL-32B based on Qwen3-VL-32B-Thinking. We release two checkpoints corresponding to different training stages: S1-VL-32B-SFT (after Stage~2) and S1-VL-32B-RL (after Stage~4). Unless otherwise specified, \textbf{S1-VL-32B} refers to \textbf{S1-VL-32B-RL} throughout this paper.

\subsection{Reasoning Mode}
\label{sec:model_reasoning}
In \textbf{Reasoning} mode, S1-VL solves the problem using structured natural-language reasoning without invoking any image manipulation tools. The input image is encoded once at the beginning, and the model completes the remaining reasoning process based on this static visual representation together with the textual context.

This mode serves as the default and foundational reasoning paradigm of S1-VL. It is suitable for a broad range of scientific tasks, including scientific question answering, multi-step derivation, and knowledge-intensive multimodal reasoning, where additional visual operations are unnecessary. Importantly, S1-VL does not treat tool use as the default behavior. When the expected benefit of image manipulation is low, the model can directly remain in Reasoning mode rather than switching to Thinking-with-Images mode. This behavior is enabled by the adaptive data routing strategy used during data construction (Section~\ref{sec:data_twi_routing}).

\subsection{Thinking-with-Images Mode}
\label{sec:model_twi}
\subsubsection{Python Code Sandbox}
\label{sec:model_twi_sandbox}
The Thinking-with-Images capability of S1-VL is implemented through the AIO Sandbox (All-in-One Agent Sandbox)~\citep{wang2025ui}, a unified execution environment built on lightweight cloud-native sandboxing technology. AIO Sandbox integrates interfaces such as Jupyter, Terminal, and the file system, and provides the model with a stateful Python execution environment for visual manipulation during reasoning.

During inference, the model generates Python code for image processing, which is executed by the Jupyter kernel inside the sandbox. The execution result, typically an intermediate image path or a computed result, is then returned to the model for the next reasoning step. A key advantage of this design is its flexibility: instead of relying on a fixed predefined toolset, the model can write arbitrary Python code to meet the needs of different reasoning scenarios. This makes it possible to support a wide range of operations, including coordinate-based cropping, local zooming, contrast or brightness enhancement, auxiliary line and bounding-box drawing, and keypoint annotation.

All turns of a single inference episode are executed within the same stateful Jupyter session, which preserves execution context across turns. At the same time, each task is assigned an independent session that is destroyed after completion, ensuring isolation between inference episodes.

\subsubsection{Prompt Template}
\label{sec:model_twi_prompt}
To make tool use reliable and reproducible, we define an explicit prompt-level interaction protocol for the Python sandbox. The system prompt specifies the functionality of the Python tool, its invocation format, and the types of image manipulation operations that are allowed during reasoning. The core content of the system prompt is shown below.

\begin{tcolorbox}[title=System Prompt, breakable, colback=gray!5]
\small
\begin{verbatim}
You are a helpful assistant.

# Tools
You may call one or more functions to assist with the user query.
You are provided with function signatures within <tools></tools> XML tags:

<tools>
{"type": "function", "function": {"name": "python", "description": "Use this tool to
execute Python code in your chain of thought.\n\nWhen you send a message containing 
Python code to python, it will be executed in a stateful Jupyter notebook environment.
python will respond with the output of the execution or time out after 60.0 seconds. 
The drive at '/mnt/data/images/temp' can be used to save the temporary image files.
Internet access for this session is disabled. Do not make external web requests or API
calls as they will fail.\n\nReasoning & Image Manipulation & Drawing Auxiliary Graphics
(Optional but Encouraged):\n- You have the capability to write executable Python code 
to perform image manipulations (e.g., cropping to a Region of Interest (ROI), resizing,
rotation, adjusting contrast) or perform calculation for better reasoning.\n- You have 
the capability to write Python code to add auxiliary graphics (such as segments, circles,
rectangles, labels, etc.) to the image, to help illustrate your reasoning process.\n- 
The code will be executed in a secure sandbox, and its output will be provided back to 
you for further analysis.\n- At the end of the code, print the path of the processed 
image (processed_path) or the relevant result for further processing within the sandbox
environment.", "parameters": {"type": "object", "properties": {"code": {"type": "string",
"description": "The Python code to execute"}}}, "required": ["code"]}}
</tools>

For each function call, return a json object with function name and arguments within 
<tool_call></tool_call> XML tags:
<tool_call>
{"name": <function-name>, "arguments": <args-json-object>}
</tool_call>
\end{verbatim}
\end{tcolorbox}
\newpage

In addition to the tool specification, we append image metadata to the user input after each image, including the image path and its dimensions. This metadata helps the model ground coordinate-based operations in the actual image space, which is particularly important for tasks involving cropping, zooming, or region localization. The corresponding user-side prompt template is shown below.

\begin{tcolorbox}[title=User Prompt, breakable, colback=gray!5]
\small
\begin{verbatim}
image path: /mnt/data/images/...
image width: <W>
image height: <H>
\end{verbatim}
\end{tcolorbox}

\subsubsection{Multi-Turn Inference Pipeline}
\label{sec:model_twi_pipeline}

\begin{figure}[t]
    \centering
    \includegraphics[width=\linewidth]{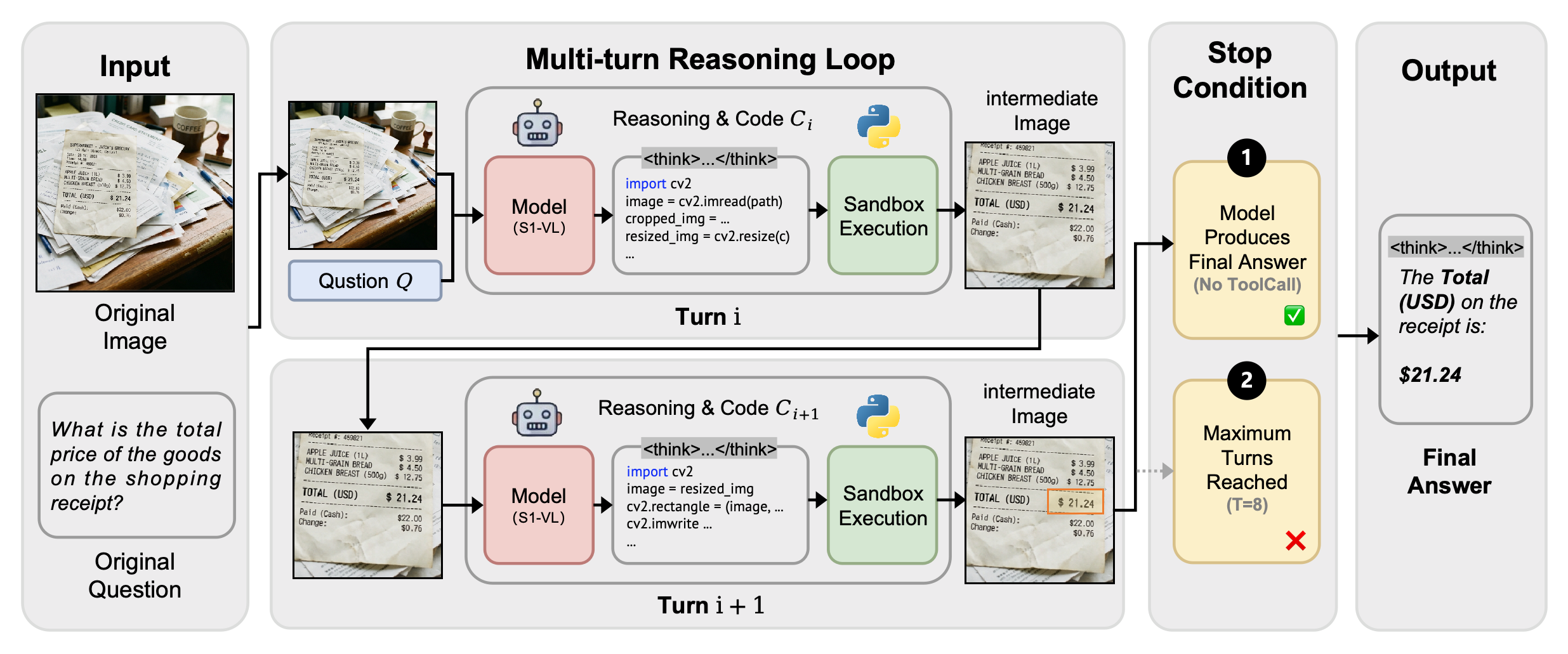}
    \caption{Overview of the Thinking-with-Images multi-turn inference pipeline in S1-VL. Given an input image and question, the model iteratively generates Python code for image manipulation, executes it in the AIO Sandbox, and incorporates the resulting intermediate image into the next reasoning turn. This process repeats until the model produces a final answer without invoking a tool call, or the maximum number of turns (8) is reached.}
    \label{fig:twi_pipeline}
\end{figure}

S1-VL performs Thinking-with-Images inference in a multi-turn closed loop. Given an input image and a question, the model may generate Python code to manipulate the image, execute the code in the sandbox, observe the returned intermediate result, and continue reasoning based on the updated visual evidence. This process repeats until the model outputs a final answer without invoking the tool, or a predefined maximum number of turns is reached. In our implementation, the maximum number of image-interaction turns is set to 8 ($\texttt{MAX\_TURNS}=8$). The overall pipeline is illustrated in Figure~\ref{fig:twi_pipeline}.

At each turn, the assistant message may contain both textual reasoning and a tool call. After execution, the returned result is injected into the next user turn as an image or execution output, thereby forming an interleaved reasoning-and-observation trajectory. All turns within the same inference episode share a single stateful Jupyter session, which ensures execution consistency across successive image operations.


\section{Data Construction}
\label{sec:data}

\begin{figure}[t]
    \centering
    \includegraphics[width=\linewidth]{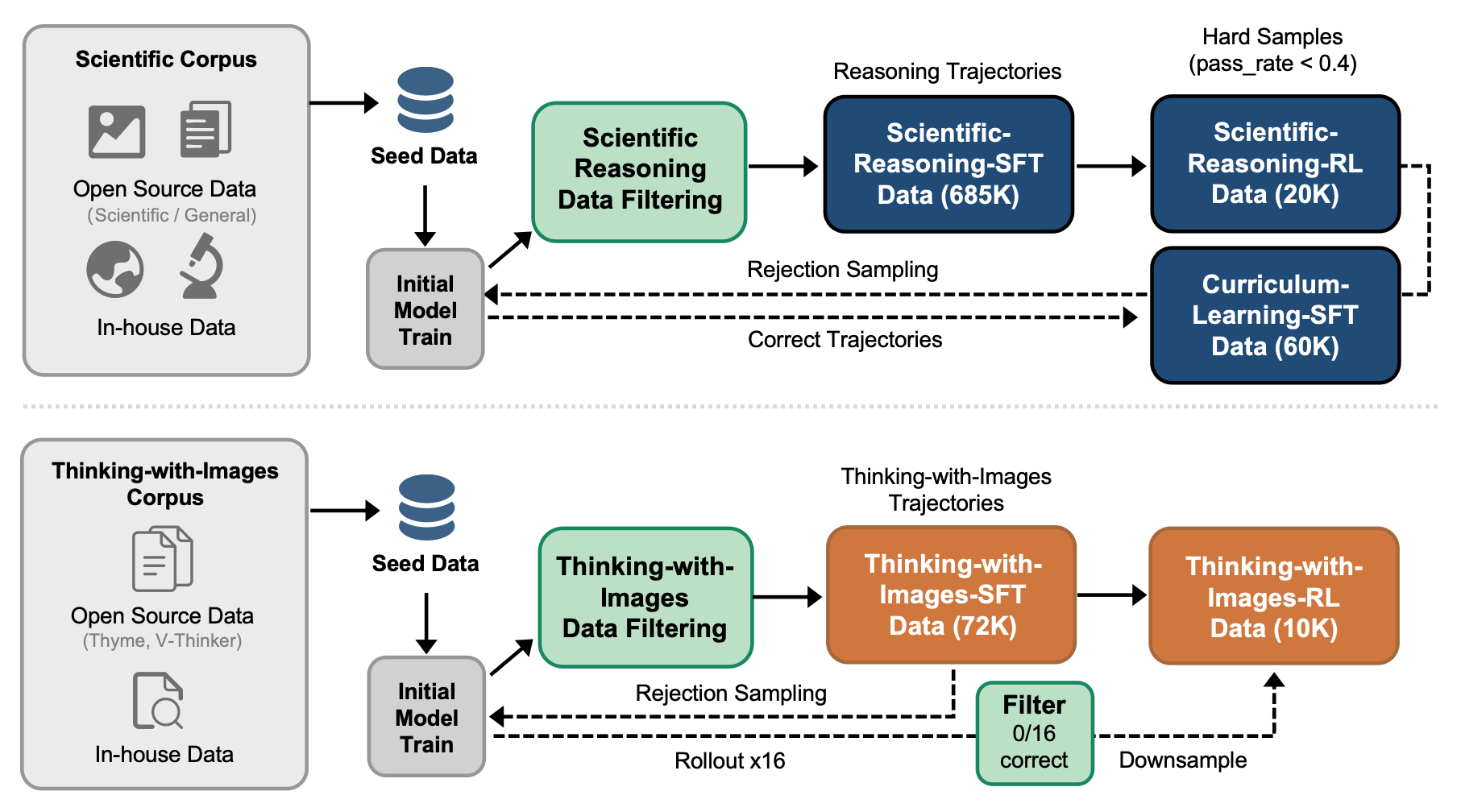}
    \caption{Overview of the two parallel data processing pipelines for scientific reasoning and Thinking-with-Images paradigms.}
    \label{fig:data_pipeline}
\end{figure}

As illustrated in Figure~\ref{fig:data_pipeline}, we build two parallel data processing pipelines, corresponding to the \textbf{Scientific Reasoning} and \textbf{Thinking-with-Images} paradigms, respectively. Both pipelines follow an iterative distillation framework consisting of four stages: seed data collection, initial model training, trajectory sampling and filtering, and final dataset construction. The scientific reasoning pipeline adopts a curriculum-based strategy to route difficult samples to the reinforcement learning stage, while the Thinking-with-Images pipeline introduces multi-dimensional quality filtering together with an adaptive data routing mechanism, enabling the model to learn when visual operations are truly necessary.

\subsection{Scientific Reasoning Data Distillation}
\label{sec:data_reasoning}

\subsubsection{Data Collection and Self-Distillation}
\label{sec:data_reasoning_collection}
We collect and curate scientific multimodal question-answering data spanning six disciplines: mathematics, physics, chemistry, astronomy, geography, and biology. The data is sourced from public datasets, scientific textbook corpus, and subject competition repositories. We first perform preliminary training on a seed dataset, and then use the trained model for self-distillation to generate chain-of-thought trajectories with complete intermediate reasoning steps. To prevent degradation in general visual understanding ability, we additionally mix a proportion of general-purpose multimodal data into the scientific corpus, yielding the \textbf{Scientific-Reasoning-SFT} dataset with 685K instances.

\subsubsection{Curriculum Learning and Difficulty Stratification}
\label{sec:data_reasoning_curriculum}
To further improve data quality and increase training data difficulty following the curriculum learning paradigm~\citep{bengio2009curriculum}, we perform multiple sampling passes for each data instance. Hard samples with $\text{pass\_rate} < 0.4$ within 10 sampling attempts are retained as training data for the reinforcement learning (RL) stage, forming the \textbf{Scientific-Reasoning-RL} dataset with 20K instances. Meanwhile, from the sampling trajectories of these data, we select 60K reasoning trajectories with correct answers to form the \textbf{Curriculum-Learning-SFT} dataset. This strategy ensures that SFT data primarily consists of high-quality correct trajectories, while RL data focuses on hard samples that the model has not yet stably mastered, with each playing a distinct role in the training pipeline.

\subsubsection{Multi-Dimensional Trajectory Filtering}
\label{sec:data_reasoning_filter}
We further apply a multi-dimensional filtering strategy to the collected correct trajectories in order to remove low-quality samples with formatting anomalies, repetitive degeneration, or reasoning artifacts. The filtering criteria are summarized as follows:

\begin{itemize}[leftmargin=*, itemsep=2pt]
    \item \textbf{Meaningless Wait Token Detection:} Remove trajectories containing excessive meaningless ``wait'' tokens, so as to prevent the model from learning ineffective placeholder-style reasoning behaviors.
    
    \item \textbf{Short Phrase Repetition Detection:} Remove degraded trajectories with repetitive short-phrase stacking, which may otherwise contaminate the training corpus with low-quality repeated content.
    
    \item \textbf{Short Phrase Multi-line Anomaly Detection:} Remove trajectories with abnormal repetitive line breaks during reasoning, which are typically caused by formatting degeneration.
    
    \item \textbf{Numeric Sequence Repetition Detection:} Remove trajectories with abnormally repetitive numeric sequence generation, so as to avoid numerical reasoning degeneration samples from disrupting training.
    
    \item \textbf{Think Format Compliance Check:} Ensure that each trajectory follows the required reasoning format, maintaining consistency between the training data and the expected output format at inference time.
    
    \item \textbf{Semantic Similarity Filtering:} Use embedding models to measure semantic similarity across reasoning steps and filter semantically redundant content.
\end{itemize}

\subsection{Thinking-with-Images Data Construction}
\label{sec:data_twi}

\subsubsection{Data Sources}
\label{sec:data_twi_source}

The Thinking-with-Images training data is constructed from two sources.

\textbf{Open-Source Data Adaptation.}
We collect open-source multi-turn tool-use trajectories from Thyme~\citep{zhang2025thyme} and V-Thinker~\citep{qiao2025v} as the initial corpus. Because the original data format differs from our training framework, we perform structured adaptation, including reconstructing assistant responses into our standard chain-of-thought and tool-calling format, injecting intermediate images from tool outputs into user-side messages in \texttt{<tool\_response>} form, and unifying all samples into a consistent multi-turn dialogue format.

\textbf{In-House High-Value Task Data.}
We also construct proprietary data targeting two categories of high-value scenarios. The first consists of real-world tasks that require fine-grained image inspection to solve effectively, such as high-resolution scientific charts and densely annotated images. The second consists of mathematical problems in which auxiliary visual operations---such as drawing lines, marking points, or placing bounding boxes---can facilitate reasoning. This second category is particularly valuable because its final answers are typically verifiable, which enables automated outcome-based quality assessment and supports the iterative training process described in Section~\ref{sec:training}.

\subsubsection{Multi-Dimensional Quality Filtering}
\label{sec:data_twi_filter}

Existing Thinking-with-Images datasets often suffer from redundant, ineffective, or erroneous visual operations, such as repeatedly manipulating the same region, generating intermediate images with little useful information, or producing mismatches between image operations and reasoning steps. To address these issues, we design a six-dimensional quality filtering framework:

\begin{itemize}[leftmargin=*, itemsep=2pt]
    \item \textbf{Format Validation.} Check the compliance and parsability of Python code blocks in each instance, and remove samples with missing or malformed tool-calling structures.
    
    \item \textbf{Reasoning-Answer Consistency Validation.} Enforce consistency between the reasoning chain and the final answer, filtering out samples whose reasoning is incorrect even if the final answer happens to be correct by chance.
    
    \item \textbf{Intermediate Image Validity Validation.} Perform lightweight validation of intermediate images produced by code execution, removing invalid outputs such as blank or solid-color images that contain no useful visual information.
    
    \item \textbf{Image-Text Consistency Validation.} Use a vision-language model to assess semantic alignment between intermediate images and the corresponding reasoning text, filtering out samples where visual operations do not match the stated reasoning intent.
    
    \item \textbf{Key Information Validity Validation.} Verify whether the final-turn image contains the core visual evidence required to solve the problem, ensuring that the overall operation chain meaningfully supports the answer.
    
    \item \textbf{Cross-Turn Redundancy Detection.} Measure the repetitiveness of visual operations across turns and remove redundant trajectories that repeatedly act on identical or highly similar regions.
\end{itemize}

\subsubsection{Adaptive Data Routing}
\label{sec:data_twi_routing}

After the above filtering stages, there remains a subset of samples whose visual operations are formally valid but do not make a substantive contribution to the final solution. Typical cases include operations applied to image regions irrelevant to the question, or trajectories in which the visual steps could be replaced entirely by direct language reasoning. Instead of discarding such samples, we apply an \textbf{adaptive data routing strategy}: we remove the image operation steps and intermediate images, retain the textual reasoning trajectory, and convert the sample into Reasoning-mode training data.

This strategy offers two key benefits. First, it avoids unnecessary data waste and improves overall data utilization. Second, it exposes the model to both kinds of supervision during training: cases where visual operations provide genuine value, and cases where direct reasoning is sufficient. As a result, the model can learn to decide at inference time whether image manipulation is actually needed, rather than treating Thinking-with-Images as the default strategy.

After filtering, we obtain the \textbf{Thinking-with-Images-SFT} dataset containing 72K instances. We then train a temporary model on the initial corpus and perform 16 rollouts for each instance in this dataset. Samples for which all 16 rollouts produce incorrect answers are removed, as they are likely to contain inherent data quality issues. From the remaining pool, we downsample 10K instances to form the \textbf{Thinking-with-Images-RL} dataset used for reinforcement learning in this paradigm.

\section{Training Pipeline}
\label{sec:training}

\begin{figure}[!t]
    \centering
    \includegraphics[width=\linewidth]{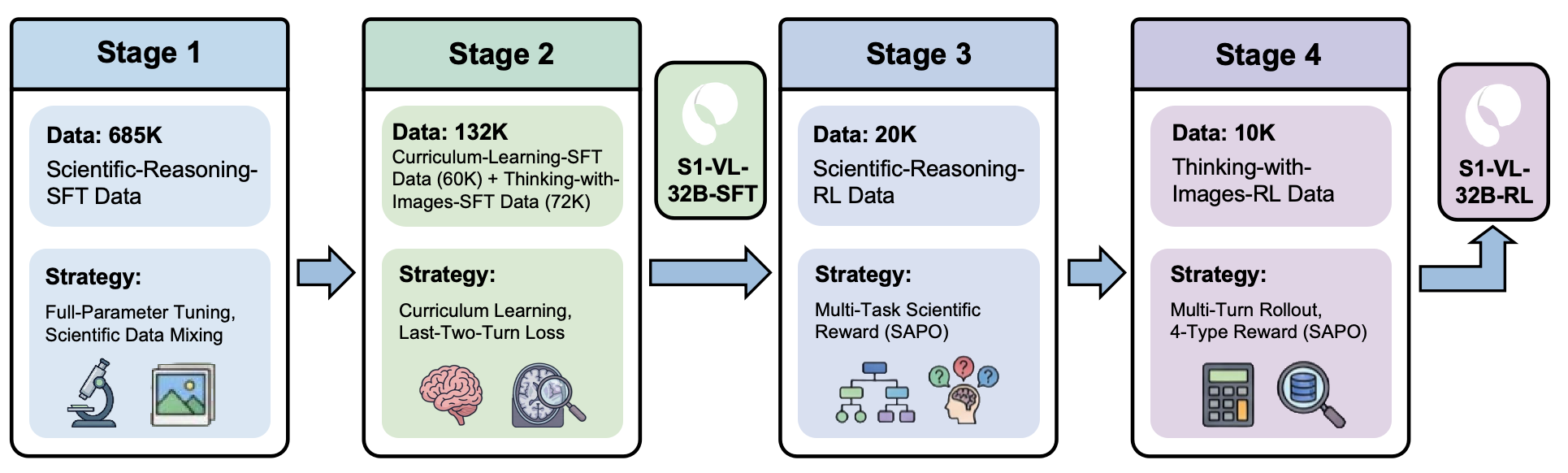}
    \caption{The four-stage progressive training pipeline of S1-VL. Each stage builds upon the previous one, progressively enhancing the model's scientific reasoning capability and Thinking-with-Images capability.}
    \label{fig:training_pipeline}
\end{figure}

S1-VL-32B is trained with a four-stage progressive pipeline. Starting from the general multimodal capabilities of Qwen3-VL-32B-Thinking, the pipeline gradually builds scientific visual understanding, strengthens domain-specific reasoning, and finally equips the model with robust Thinking-with-Images capability. Specifically, the four stages are: (1) general scientific multimodal supervised fine-tuning, (2) Thinking-with-Images cold-start fine-tuning augmented with curated scientific hard examples, (3) SAPO-based scientific reinforcement learning, and (4) SAPO-based Thinking-with-Images reinforcement learning. The overall training process is illustrated in Figure~\ref{fig:training_pipeline}.

\subsection{Stage 1: Scientific Reasoning SFT}
\label{sec:training_stage1}

Stage 1 establishes the model's initial foundation in scientific visual understanding and reasoning on top of Qwen3-VL-32B-Thinking. The training data is primarily composed of scientific multimodal instruction data spanning six disciplines: mathematics, physics, chemistry, astronomy, geography, and biology. To avoid degrading the model's general visual understanding ability as it becomes more specialized toward scientific domains, we additionally mix in a proportion of general-purpose multimodal data. The resulting training set contains 685K instances in total. This stage is trained with full-parameter supervised fine-tuning.

\subsection{Stage 2: Thinking-with-Images SFT}
\label{sec:training_stage2}

Stage 2 builds on Stage 1 by introducing the Thinking-with-Images paradigm while continuing to strengthen scientific-domain reasoning. The goal of this stage is to equip the model with an initial ability to invoke Python-based image operations during reasoning, without sacrificing its scientific reasoning foundation.

The training data in this stage contains 132K instances from two sources: 60K high-quality reasoning trajectories curated from hard scientific samples (the \textbf{Curriculum-Learning-SFT} dataset in Section~\ref{sec:data_reasoning_curriculum}) and 72K Thinking-with-Images instances obtained through the filtering pipeline described in Section~\ref{sec:data_twi}. The former reinforces scientific reasoning, while the latter introduces supervised multi-turn image interaction. These two components are mixed for joint training.

In multi-turn Thinking-with-Images supervision, we find that assigning equal loss to all turns introduces substantial noise from early exploratory interactions. Early image-operation steps are often tentative and relatively low-quality, while the final stages of the trajectory are more directly related to answer correctness and thus provide stronger supervision.

We therefore adopt a \textbf{Last-Two-Turn Loss} strategy: for each trajectory, loss is computed only on the last tool-use turn and the final answer turn, while all earlier turns are ignored. The former captures the final image-manipulation decision, and the latter captures the final pure-text reasoning and answer generation. This design focuses learning on the most informative part of the trajectory and reduces interference from noisy early exploration. Empirically, it improves both code generation quality and final task performance.

\subsection{Stage 3: Scientific Reasoning RL}
\label{sec:training_stage3}

Stage 3 further improves the model's performance on challenging scientific reasoning tasks through reinforcement learning. Starting from the Stage 2 checkpoint, this stage focuses on hard scientific problems for which supervised fine-tuning alone is insufficient to achieve stable reasoning performance.

\textbf{Training Data.}
The training data for this stage consists of hard samples selected through the rejection-sampling procedure described in Section~\ref{sec:data_reasoning_curriculum}, namely scientific multimodal reasoning problems whose pass rate is below 0.4 over 10 sampling attempts. These samples remain sufficiently challenging for the model and therefore provide meaningful exploration space during RL training.

\textbf{Multi-Task Scientific Reward Function.}
Scientific reasoning tasks vary substantially across disciplines, and their outputs cannot be reliably evaluated using a single uniform criterion. Different tasks impose different requirements on answer format, numerical validity, unit consistency, derivation logic, and semantic correctness. To accommodate this heterogeneity, we design a unified yet task-aware reward framework that combines format checking, rule-based verification, and model-based semantic judgment.

Formally, let $\mathcal{T}$ denote the set of task types, such as mathematics, physics, astronomy, and chemistry. Given an input $x$ and a model output $y$, we use the indicator function $\mathbf{1}(t|x)$ to identify the task type of $x$. The overall reward $R(x, y)$ is defined as a task-conditioned weighted combination of three complementary components:

\begin{equation}
    R(x, y) = \sum_{t \in \mathcal{T}} \mathbf{1}(t|x) 
    \left[ 
        \lambda_f^{(t)} R_{\text{format}}(y) + 
        \lambda_r^{(t)} R_{\text{rule}}^{(t)}(y) + 
        \lambda_j^{(t)} R_{\text{judge}}^{(t)}(x, y) 
    \right]
    \label{eq:overall_reward}
\end{equation}

The first component, $R_{\text{format}}(y)$, evaluates whether the output adheres to the required structural and formatting constraints (e.g., boxed answers, unit specifications). The second component, $R_{\text{rule}}^{(t)}(y)$, measures task-specific correctness through rule-based or symbolic verification, and can be further decomposed into multiple sub-rewards:

\begin{equation}
    R_{\text{rule}}^{(t)}(y) = \sum_{k} \alpha_k^{(t)} R_{\text{rule},k}^{(t)}(y)
    \label{eq:rule_reward}
\end{equation}

\noindent where each $R_{\text{rule},k}^{(t)}$ captures a distinct verifiable 
dimension of correctness (e.g., numerical accuracy, unit consistency, or logical 
validity) with corresponding weight $\alpha_k^{(t)}$. The third component, 
$R_{\text{judge}}^{(t)}(x, y)$, provides a complementary semantic evaluation via an 
LLM-based judge, defined as the expected score over a set of judges $\mathcal{J}$:

\begin{equation}
    R_{\text{judge}}^{(t)}(x, y) = \mathbb{E}_{j \in \mathcal{J}}\left[ s_j(x, y) \right]
    \label{eq:judge_reward}
\end{equation}

This design is motivated by the observation that rule-based verification alone is often insufficient for open-ended scientific tasks, where partial credit, alternative 
solution paths, or free-form reasoning must also be assessed. The task-dependent 
coefficients $\lambda_f^{(t)}$, $\lambda_r^{(t)}$, and $\lambda_j^{(t)}$ allow 
flexible calibration of each component's contribution across different disciplines, 
ensuring that the reward signal remains both domain-aware and evaluation-robust.

To optimize the proposed multi-task reward function, we adopt the 
\textbf{SAPO} algorithm. By substituting our multi-task scientific reward $R(x,y)$ into the SAPO framework, we construct the full \textbf{Scientific-RL} training pipeline. For each training sample $x$, the task type $t$ is first identified via $\mathbf{1}(t|x)$, and the corresponding reward components  $R_{\text{format}}$, $R_{\text{rule}}^{(t)}$, and $R_{\text{judge}}^{(t)}$ 
are computed and aggregated according to Eq.~\eqref{eq:overall_reward}. 
The resulting scalar reward is then used by SAPO to update the policy, 
enabling the model to receive dense, domain-aware supervision across 
diverse scientific disciplines within a single unified training process.

\subsection{Stage 4: Thinking-with-Images RL}
\label{sec:training_stage4}

Stage 4 builds upon Stage 3 to further enhance the model's Thinking-with-Images capability through reinforcement learning, enabling the model to stably generate high-quality reasoning and tool-calling sequences in complex multi-turn image operation scenarios.

\subsubsection{Rollout Pipeline}
\label{sec:training_stage4_rollout}

RL training in multi-turn tool-calling scenarios faces unique engineering challenges: the rollout process for each training sample is no longer pure text generation, but requires actually executing Python code, invoking the AIO Sandbox, obtaining intermediate images, and injecting these images into subsequent reasoning steps during generation, forming a complete multi-turn interaction chain.

To ensure the stable operation of the rollout process, we construct a dedicated multi-turn tool-calling rollout pipeline with the following core design principles. First, each sample's rollout independently creates an AIO Sandbox Jupyter session, ensuring complete isolation of execution environments across different samples and preventing state contamination. Second, a multi-turn planner is designed to manage the state of tool-calling results at each turn, handling exceptions such as code execution failures and timeouts to guarantee the robustness of the entire rollout chain. Third, validity verification is performed on code execution results to filter out invalid intermediate images such as blank or solid-color images, injecting only intermediate images containing substantive visual content into subsequent reasoning steps.

\subsubsection{Thinking-with-Images Reward}
\label{sec:training_stage4_reward}

The reward function in this stage comprises four types of reward signals covering quality assessment across the entire tool-calling pipeline, as detailed in Table~\ref{tab:reward}.

\begin{table}[t]
    \centering
    \caption{Four reward signals used in the reinforcement learning training stage.}
    \label{tab:reward}
    \small
    \begin{tabular*}{\textwidth}{@{} p{0.2\textwidth} @{\hspace{1.5em}} p{0.75\textwidth} @{}}
        \toprule
        \textbf{Reward Type} & \textbf{Description} \\
        \midrule
        Format Reward & Evaluates whether the generated output adheres to the required structured paradigm, \emph{i.e.}, \texttt{<think>...</think>}\texttt{\textbackslash n}\texttt{<tool\_call>...</tool\_call>}; penalizes any deviation from the prescribed format. \\
        \midrule
        Answer Reward & Evaluates the consistency between the final answer and the verifiable ground-truth answer; serves as the core reward signal. \\
        \midrule
        Consistency Reward & Comprises two complementary sub-rewards. \textbf{(1)~Think--Answer Consistency}: an LLM Judge assesses whether the reasoning content within \texttt{<think>} logically entails the final answer following \texttt{</think>}. \textbf{(2)~Crop Correctness}: evaluates whether the last cropped image, combined with the question, is sufficient to derive the correct answer. \\
        \midrule
        Tool-Calling Efficiency Reward & Assigns full score when the tool is invoked at most once; applies cosine decay to penalize redundant multiple invocations, encouraging concise and efficient tool use. \\
        \bottomrule
    \end{tabular*}
\end{table}

To guide the model toward accurate, well-structured, and efficient visual reasoning,
we design a composite reward function comprising four complementary signals:
format reward $r_{\text{fmt}}$, answer correctness reward $r_{\text{acc}}$,
consistency reward $r_{\text{con}}$, and tool-calling efficiency reward $r_{\text{tool}}$.

\paragraph{Format Reward.}
The format reward $r_{\text{fmt}}$ acts as a hard gate on output structure.
If the generated response does not conform to the required paradigm—
\texttt{<think>...</think>}\textbackslash n\texttt{<tool\_call>...</tool\_call>}—
the entire reward is overridden and set to $-1$, regardless of other signals.

\paragraph{Answer Correctness Reward.}
The answer correctness reward $r_{\text{acc}} \in \{0, 1\}$ evaluates whether the
model's final answer matches the verifiable ground-truth.
A mismatch yields $r_{\text{acc}} = 0$, which further suppresses the overall reward
to $0$ as a secondary hard constraint. We assign the largest weight (0.5) to this term to ensure that task correctness remains the dominant learning signal. This prevents the model from exploiting auxiliary rewards (e.g., tool usage or formatting) without solving the underlying task.

\paragraph{Consistency Reward.}
The consistency reward $r_{\text{con}} \in [0, 1]$ comprises two sub-components:

\begin{itemize}[leftmargin=*, itemsep=2pt]
    \item \textbf{Think--Answer Consistency} ($r_{\text{think}}$): An LLM Judge
    evaluates whether the chain-of-thought content within \texttt{<think>} logically
    entails the final answer appearing after \texttt{</think>}.

    \item \textbf{Crop Correctness} ($r_{\text{crop}}$): Given the original question
    and the last cropped image produced by the tool call, a judge assesses whether
    the cropped region contains sufficient visual information to derive the correct answer.
\end{itemize}

\noindent The two sub-scores are averaged to form the overall consistency reward:
\begin{equation}
    r_{\text{con}} = \frac{1}{2}\left(r_{\text{think}} + r_{\text{crop}}\right)
\end{equation}
Importantly, this component incorporates \textbf{visual grounding verification} when tool-generated images are available, ensuring that the reasoning is consistent with the cropped or processed visual evidence. This term is weighted at 0.3 to encourage coherent reasoning, faithful use of visual information, and reduced hallucination.

\paragraph{Tool-Calling Efficiency Reward.}
To discourage redundant tool invocations, the efficiency reward $r_{\text{tool}} \in [0, 1]$
is defined as:
\begin{equation}
    r_{\text{tool}} =
    \begin{cases}
        0, & \text{if } n = 0 \\[4pt]
        1, & \text{if } n = 1 \\[4pt]
        \dfrac{1}{2}\left(1 + \cos\!\left(\dfrac{(n-1)\,\pi}{N_{\max} - 1}\right)\right),
        & \text{if } 1 < n \leq N_{\max}
    \end{cases}
\end{equation}
where $n$ denotes the number of tool-call invocations and $N_{\max} = 8$ is the
maximum allowed number of invocations per trajectory. The efficiency reward $r_{\text{tool}}$ encourages economical use of tools by penalizing unnecessary or excessive tool calls. Specifically, zero tool usage receives a low score, a single tool call achieves the highest score, and multiple calls are softly penalized via a decay function. We assign a relatively small weight (0.1) to this term to avoid over-regularizing the policy. This ensures that the model is not discouraged from using tools when necessary, while still promoting concise interaction patterns.

\paragraph{Tool Usage Bonus.}
The tool usage bonus $r_{\text{bonus}}$ provides an additional positive signal when the model invokes tools during reasoning:
\begin{equation}
r_{\text{bonus}} =
\begin{cases}
1, & \text{if at least one tool call is used} \\
0, & \text{otherwise}
\end{cases}
\end{equation}

This term encourages early-stage exploration of tool usage and helps establish a causal link between tool invocation and improved performance. It prevents the model from collapsing into a degenerate policy that avoids tool interaction entirely.

\paragraph{Overall Reward.}
The overall reward design reflects a balance between competing objectives: correctness (0.5), reasoning quality (0.3), and tool efficiency (0.1), along with an auxiliary exploration bonus (0.1). Importantly, the relatively low weight on efficiency avoids introducing a global bias against tool usage, while the consistency and bonus terms implicitly encourage \textbf{tool usage only when it improves reasoning and accuracy}.

\begin{equation}
    \mathcal{R} =
    \begin{cases}
        -1, & \text{if } r_{\text{fmt}} = 0 \\[4pt]
        0.5 \cdot r_{\text{acc}} + 0.3 \cdot r_{\text{con}} + 0.1 \cdot r_{\text{tool}} + 0.1 \cdot r_{\text{bonus}},
        & \text{otherwise}
    \end{cases}
    \label{eq:overall_reward_final}
\end{equation}

\noindent We adopt the \textbf{SAPO} to optimize the model under
the above reward scheme. Concretely, the policy $\pi_\theta$ is trained to maximize the expected reward $\mathbb{E}_{\tau \sim \pi_\theta}[\mathcal{R}(\tau)]$ over sampled trajectories
$\tau$. The maximum number of tool-call rounds per trajectory is set to
$N_{\max} = 8$, allowing the model to perform iterative visual inspection
while being discouraged from excessive invocations via $r_{\text{tool}}$.

\paragraph{Reward Hacking Analysis.}

During the training process, we identified a critical instance of reward hacking stemming from a misaligned reward formulation. The original reward function was 
defined as:

\begin{equation}
    R = 0.5 \cdot r_{\text{acc}} + 0.3 \cdot r_{\text{con}} + 0.2 \cdot r_{\text{tool}}
\end{equation}

\noindent where $r_{\text{tool}}$ was designed to penalize excessive tool 
invocations: the model received a full efficiency score for invoking zero or one tool calls, with a decaying penalty applied for two or more invocations.

This formulation inadvertently incentivized the model to avoid tool use entirely. Since zero tool invocations yielded a perfect $r_{\text{tool}}$ score, the model had no intrinsic motivation to engage with external tools. Furthermore, during early training, tool-free direct inference proved inherently more stable: invoking a tool introduced an additional verification step—checking whether the cropped image was correctly processed—whereas pure reasoning bypassed this step altogether, making it easier to achieve a high $r_{\text{con}}$ score. Consequently, the policy converged to a degenerate solution of \emph{never} invoking any tool, undermining the intended behavior.

To address this reward hacking, we arrived at the final reward formulation in Eq.~\eqref{eq:overall_reward_final}. Two key modifications were introduced. First, we incorporated an explicit \textbf{tool bonus} term $r_{\text{bonus}}$, which assigns a full score whenever at least one tool is invoked and zero otherwise. This directly encourages tool engagement from the early stages of training. Second, we revised the $r_{\text{tool}}$ metric: zero tool invocations now yields a score of $0$, a single invocation yields a full score, and two or more invocations incur a decaying penalty. By eliminating the reward for tool-free behavior, the revised formulation removes the perverse incentive that caused the original hacking.

\begin{figure}[t]
    \centering
    \begin{subfigure}[b]{0.48\textwidth}
        \centering
        \includegraphics[width=\textwidth]{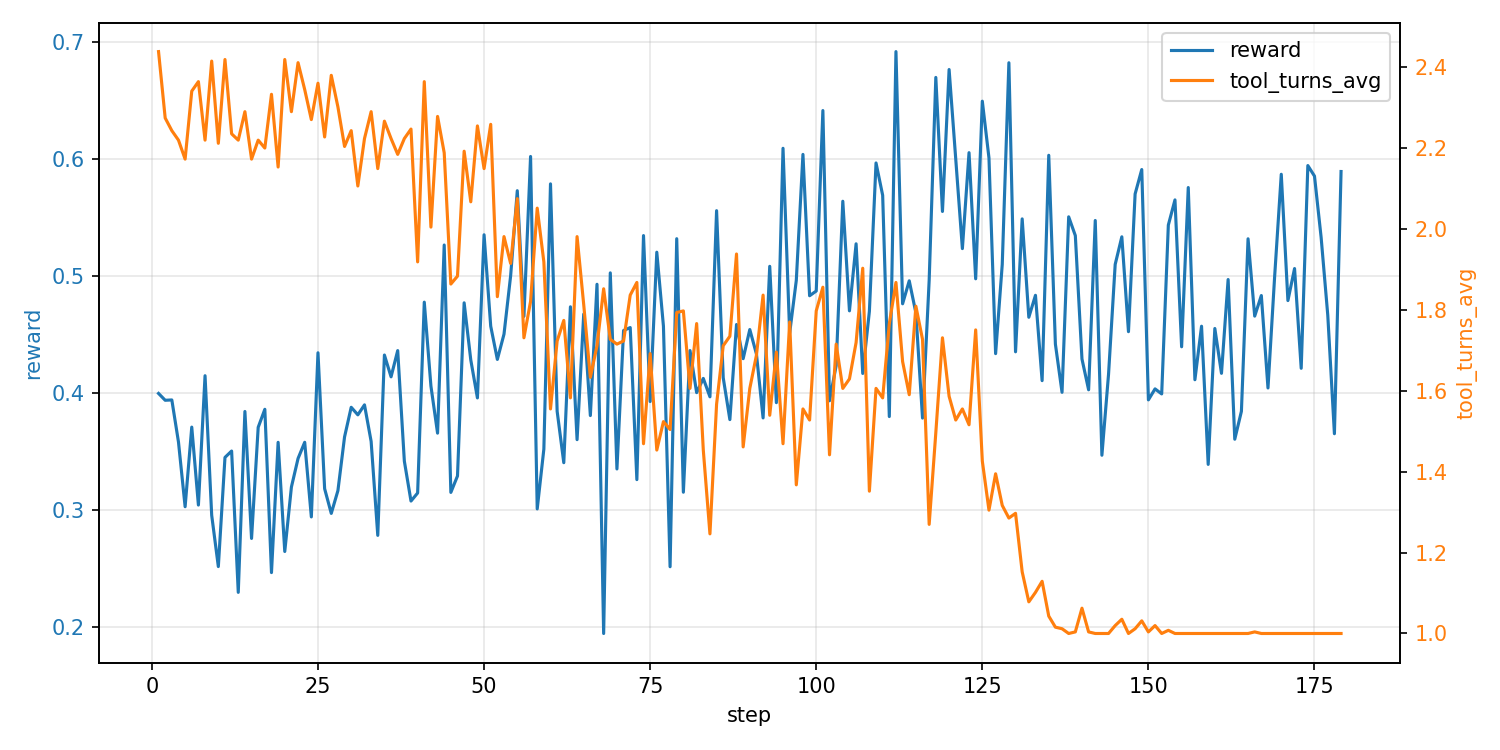}
        \caption{Reward Hacking}
        \label{fig:reward_hack}
    \end{subfigure}
    \hfill
    \begin{subfigure}[b]{0.48\textwidth}
        \centering
        \includegraphics[width=\textwidth]{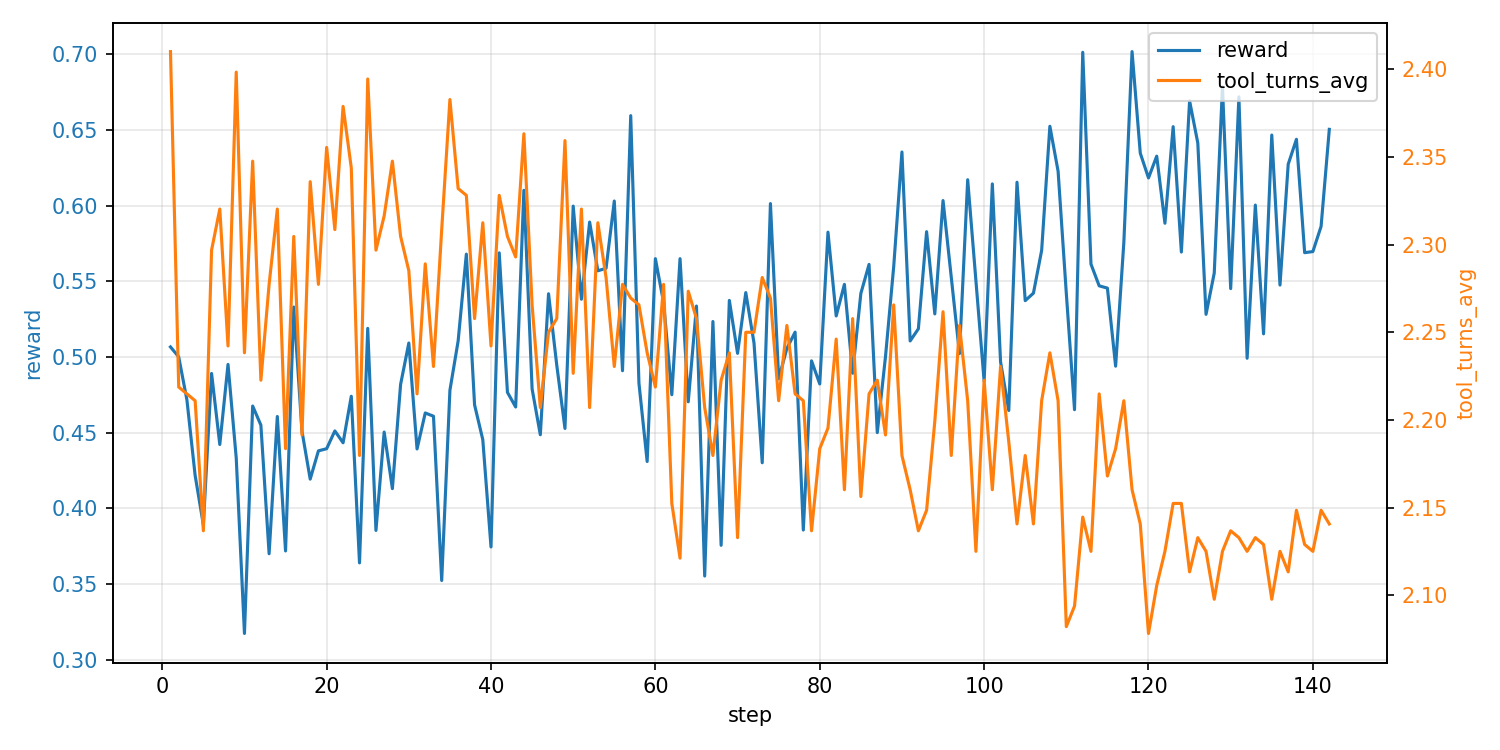}
        \caption{Correct Reward Function}
        \label{fig:correct_reward}
    \end{subfigure}
    \caption{Comparison of reward trajectories before and after reward function revision.}
    \label{fig:comparison}
\end{figure}

The revised reward function successfully eliminates the observed hacking behavior. As illustrated in Figures~\ref{fig:reward_hack} and~\ref{fig:correct_reward}, the updated formulation 
yields more stable reward trajectories throughout training, confirming that the policy no longer collapses to the degenerate no-tool solution.

\subsection{Training Hyperparameters}
\label{sec:training_hyper}

We train S1-VL using the \texttt{ms-swift} framework~\citep{zhao2024swiftascalablelightweightinfrastructure}, which provides unified support for multimodal supervised fine-tuning and reinforcement learning. All experiments are conducted on a GPU cluster with up to 8 nodes, each equipped with 8 NVIDIA A100 GPUs (80GB memory per GPU), interconnected via InfiniBand (IB). To support full-parameter training of the 32B model under long-context multimodal settings, we adopt DeepSpeed ZeRO-3~\citep{rasley2020deepspeed} together with sequence parallelism for memory optimization and distributed training efficiency.

The key stage-wise training hyperparameters are summarized in Table~\ref{tab:hyperparams}.

\begin{table}[t]
    \centering
    \caption{Training hyperparameters for each stage of the S1-VL-32B training pipeline.}
    \label{tab:hyperparams}
    \small
    \begin{tabular*}{\textwidth}{l @{\hspace{2em}\extracolsep{\fill}} cccc}
        \toprule
        \textbf{Hyperparameter} & \textbf{Stage 1} & \textbf{Stage 2} & \textbf{Stage 3} & \textbf{Stage 4} \\
        \midrule
        Training Algorithm    & SFT  & SFT  & SAPO  & SAPO  \\
        Learning Rate         & 1e-5 & 1e-5 & 1e-6  & 1e-6  \\
        Global Batch Size     & 64   & 64   & 256   & 256   \\
        Max Sequence Length   & 32768 & 32768 & 32768 & 32768 \\
        \bottomrule
    \end{tabular*}
\end{table}

\section{Experiments}
\label{sec:exp}

We release two checkpoints: S1-VL-32B-SFT (after supervised fine-tuning) and S1-VL-32B-RL (after reinforcement learning). Unless otherwise specified, S1-VL-32B refers to S1-VL-32B-RL throughout this paper.

\subsection{Evaluation Benchmarks}
\label{sec:exp_bench}

We conduct systematic evaluations of S1-VL-32B on 13 benchmarks, covering two complementary aspects: scientific reasoning and Thinking-with-Images.

\textbf{Scientific Reasoning Benchmarks.}
The scientific reasoning setting includes eight benchmarks spanning diverse scientific disciplines and multimodal reasoning types, as summarized in Table~\ref{tab:sci_benchmarks}. Collectively, they evaluate whether the model can integrate visual observations with domain knowledge, perform structured reasoning over scientific evidence, and produce accurate judgments in specialized contexts. Such benchmarks are therefore well suited for assessing the model’s capacity for generalizable multimodal scientific reasoning across heterogeneous scientific tasks.

\begin{table}[t]
    \centering
    \caption{Scientific reasoning benchmarks used for evaluation, covering eight disciplines and reasoning types.}
    \label{tab:sci_benchmarks}
    \small
    \begin{tabular*}{\textwidth}{l @{\hspace{2em}\extracolsep{\fill}} p{0.60\textwidth}}
        \toprule
        \textbf{Benchmark} & \textbf{Focus} \\
        \midrule
        MMMU~\citep{yue2024mmmu}             & Multi-disciplinary multimodal understanding \& reasoning \\
        SFE~\citep{zhou2025scientists}              & Scientific chart understanding \& information extraction \\
        MathVision\citep{wang2024measuring}       & Mathematical visual reasoning \\
        Physics~\citep{feng2025physics}          & Physics scene multimodal reasoning \\
        ScienceOlympiad~\citep{bytedance_seed_2025_scienceolympiad}  & Scientific competition reasoning \\
        VRSBench~\citep{li2024vrsbench}    & Remote sensing image understanding \& analysis \\
        GMAI-MMBench~\citep{chen2024gmai}     & Medical multimodal understanding \\
        Galaxy-10-DECaLS~\citep{Galaxy10} & Astronomical image classification \& understanding \\
        \bottomrule
    \end{tabular*}
\end{table}

\textbf{Thinking-with-Images Benchmarks.}
The Thinking-with-Images setting includes five benchmarks that emphasize fine-grained visual understanding in high-resolution images, real-world scenes, and visual search tasks, as summarized in Table~\ref{tab:twi_benchmarks}. These benchmarks place strong demands on local detail perception and are therefore particularly suitable for evaluating Thinking-with-Images capability. On such tasks, the model’s ability to actively acquire critical local evidence through image manipulation often directly determines final performance.

\begin{table}[t]
    \centering
    \caption{Thinking-with-Images benchmarks used for evaluation, focusing on high-resolution visual understanding and real-world scene comprehension.}
    \label{tab:twi_benchmarks}
    \small
    \begin{tabular*}{\textwidth}{l @{\hspace{2em}\extracolsep{\fill}} p{0.60\textwidth}}
        \toprule
        \textbf{Benchmark} & \textbf{Focus} \\
        \midrule
        HRBench-4K\citep{hrbench}          & Fine-grained visual understanding of 4K resolution images \\
        HRBench-8K\citep{hrbench}          & Fine-grained visual understanding of 8K resolution images \\
        MME-RealWorld-CN\citep{zhang2024mme}    & Real-world multimodal reasoning (Chinese) \\
        MME-RealWorld-Lite\citep{zhang2024mme}  & Real-world multimodal reasoning \\
        V*\citep{vstar}                  & Visual search and fine-grained object localization \\
        \bottomrule
    \end{tabular*}
\end{table}

\subsection{Baselines}
\label{sec:exp_baseline}

We select the following three categories of comparison models as baselines.

\textbf{Proprietary General-Purpose Models.}
The first category consists of proprietary large models, including GPT-5~\citep{GPT-5} and the Gemini 2.5~\citep{comanici2025gemini} series (Flash and Pro). These models represent the current state of the art among commercial closed-source multimodal large models, with parameter counts far exceeding that of S1-VL-32B. Their inclusion aims to validate the competitiveness of S1-VL-32B in scientific reasoning scenarios.

\textbf{Open-Source Multimodal Models.}
The second category comprises open-source multimodal models, including Qwen3-VL-32B-Thinking, Qwen3-VL-235B-A22B-Thinking~\citep{bai2025qwen3}, InternVL3-8B, and the Intern-S1 series~\citep{bai2025intern} (including Intern-S1, a large-scale scientific multimodal reasoning model built on a 235B MoE language model with a 6B vision encoder, and Intern-S1-mini, a lightweight counterpart built on an 8B dense language model with a 0.3B vision encoder). Among these, Qwen3-VL-32B serves as the base model of S1-VL-32B, and the comparison between the two directly reflects the extent to which our training pipeline improves upon the base model's capabilities.

\textbf{Thinking-with-Images Models.}
The third category consists of Thinking-with-Images specialist models, including Thyme-VL (7B)~\citep{zhang2025thyme} and Skywork-R1V4-30B (A3B)~\citep{zhang2025skywork}. These models are most closely aligned with S1-VL-32B in terms of reasoning paradigm and serve as the most direct points of comparison. It should be noted that Skywork-R1V4-30B does not provide an official inference interface, and results on some scientific reasoning benchmarks are therefore unavailable.

\begin{table}[t]
    \centering
    \caption{Scientific reasoning results on eight benchmarks. \textbf{Bold} denotes the best performance among all models.}
    \label{tab:sci_results}
    \small
    \begin{adjustbox}{max width=\textwidth}
    \begin{tabular}{lcccccccc}
        \toprule
        \textbf{Model} & \textbf{MMMU} & \textbf{SFE} & \textbf{MathVision} & \textbf{Physics} & \textbf{ScienceOlympiad} & \textbf{VRSBench} & \textbf{GMAI} & \textbf{Galaxy-10} \\
        \midrule
        Qwen3-VL-32B-Thinking       & 76.00 & 37.50 & 71.51 & 41.71 & 22.35 & 68.41 & 53.36 & 30.44 \\
        Qwen3-VL-235B-A22B-Thinking & 77.89 & 39.98 & 74.87 & 46.03 & 36.47 & 68.94 & 55.91 & 33.93 \\
        Intern-S1-mini (8B)         & 70.00 & 36.93 & 52.47 & 30.53 & 11.82 & 54.89 & 50.57 & 39.06 \\
        Intern-S1 (235B+6B)         & 75.56 & 43.98 & 63.03 & 44.95 & 36.47 & 63.48 & 58.44 & 48.37 \\
        Thyme-VL (7B)               & 48.66 & 25.37 & 26.28 & 12.41 &  1.97 & 57.61 & 42.61 & 27.84 \\
        Gemini 2.5 Flash            & 72.70 & 37.60 & 70.10 & 37.10 & 28.57 & 64.23 & 56.81 & 15.78 \\
        Gemini 2.5 Pro              & 82.00 & 43.00 & 73.30 & 40.00 & 24.13 & 65.70 & 60.32 & 25.14 \\
        GPT-5                       & 81.22 & \textbf{44.06} & 75.66 & 48.34 & \textbf{41.38} & 65.89 & 56.88 & \textbf{57.72} \\
        \midrule
        S1-VL-32B-SFT                   & 82.50 & 42.58 & 75.89 & 43.80 & 33.00 & 72.34 & 58.40 & 42.45  \\
        S1-VL-32B-RL                   & \textbf{83.40}  & 43.10  & \textbf{77.70}  & \textbf{54.35}  & 33.50  & \textbf{74.32}  & \textbf{62.13}  & 51.52  \\
        
        \bottomrule
    \end{tabular}
    \end{adjustbox}
\end{table}

\subsection{Scientific Reasoning Results}
\label{sec:exp_science}

The complete scientific reasoning evaluation results of S1-VL-32B on eight benchmarks are presented in Table~\ref{tab:sci_results}. Overall, S1-VL-32B consistently outperforms the base model Qwen3-VL-32B-Thinking across all eight benchmarks, validating the effectiveness of our training pipeline. More notably, S1-VL-32B surpasses models with substantially larger parameter counts on multiple benchmarks: it outperforms Qwen3-VL-235B-A22B-Thinking, which has approximately $7\times$ the parameters, on MMMU, MathVision, Physics, VRSBench, and GMAI, and even exceeds proprietary models such as GPT-5 and Gemini 2.5 Pro on Physics and VRSBench. The advantage on Physics is particularly striking, where S1-VL-32B achieves 54.35, surpassing GPT-5 (48.34) by 6.01 points and Qwen3-VL-235B-A22B-Thinking (46.03) by 8.32 points, demonstrating strong capability in physics-scene visual reasoning.

Compared with Intern-S1 (235B+6B), another scientific multimodal reasoning model with a significantly larger backbone, S1-VL-32B achieves superior results on MMMU, MathVision, Physics, and VRSBench, indicating that a well-designed scientific training pipeline combined with reinforcement learning can effectively compensate for the gap in parameter scale. These results confirm that S1-VL-32B establishes a strong baseline for scientific multimodal reasoning at the 32B scale.

\subsection{Thinking-with-Images Results}
\label{sec:exp_twi}

The complete Thinking-with-Images evaluation results on five benchmarks are presented in Table~\ref{tab:twi_results}. Both released checkpoints---S1-VL-32B-SFT and S1-VL-32B-RL---comprehensively surpass all compared models, demonstrating the systematic advantage of our data construction and training pipeline. S1-VL-32B-SFT already exhibits strong performance, outperforming the previously best Thinking-with-Images specialist model Skywork-R1V4-30B as well as proprietary models such as Gemini 2.5 Pro across all five benchmarks, validating the effectiveness of our multi-dimensional quality filtering and adaptive data routing strategies at the SFT stage.

Building on this foundation, S1-VL-32B-RL further achieves substantial improvements through Stage~4 Thinking-with-Images reinforcement learning, attaining state-of-the-art results on all five benchmarks. The gains are particularly pronounced in the high-resolution understanding direction: HRBench-4K improves from 85.00 (SFT) to 91.38 (RL), and HRBench-8K from 85.10 to 93.50, surpassing the previous best results (Gemini 2.5 Pro, 83.90 and 81.50) by 9.10 and 11.00 points respectively. On V*, S1-VL-32B-RL achieves 92.70, exceeding Skywork-R1V4-30B (88.00) by 4.70 points. The consistent improvement from SFT to RL further validates the effectiveness of our composite reward design and the SAPO algorithm in the Thinking-with-Images setting.

\begin{table}[t]
    \centering
    \caption{Thinking-with-Images results on five benchmarks. \textbf{Bold} denotes the best performance. S1-VL-32B achieves state-of-the-art across all five benchmarks.}
    \label{tab:twi_results}
    \small
    \begin{tabular*}{\textwidth}{l @{\hspace{1.5em}\extracolsep{\fill}} ccccc}
        \toprule
        \textbf{Model} & \textbf{HRBench-4K} & \textbf{HRBench-8K} & \textbf{MME-RW-CN} & \textbf{MME-RW-Lite} & \textbf{V*} \\
        \midrule
        \multicolumn{6}{c}{\cellcolor{gray!12}\textit{Open-Source Models}} \\
        Qwen2.5-VL-7B               & 68.80 & 65.30 & 60.80 & 44.10 & 76.40 \\
        Qwen2.5-VL-32B              & 73.40 & 70.40 & 60.50 & 46.20 & 81.20 \\
        Qwen3-VL-32B-Thinking       & 82.63 & 77.00 & 61.21 & 47.94 & 86.91 \\
        Qwen3-VL-235B-A22B-Thinking & 83.00 & 80.40 & 68.80 & 56.50 & 80.60 \\
        InternVL3-8B                & 70.00 & 69.30 & 60.50 & 48.60 & 70.20 \\
        Intern-S1-mini (8B)         & 62.13 & 54.38 & 54.67 & 40.02 & 58.64 \\
        Intern-S1 (235B+6B)         & 82.50 & 66.38 & 62.98 & 56.48 & 74.86 \\
        \multicolumn{6}{c}{\cellcolor{gray!12}\textit{Proprietary Models}} \\
        Gemini 2.5 Flash            & 77.50 & 73.70 & 61.20 & 50.20 & 72.30 \\
        Gemini 2.5 Pro              & 83.90 & 81.50 & 69.30 & 58.30 & 79.10 \\
        GPT-5                       & 74.25 & 73.75 & 63.97 & 57.22 & 74.87 \\
        \multicolumn{6}{c}{\cellcolor{gray!12}\textit{Thinking-with-Images Specialist Models}} \\
        Thyme-VL (7B)               & 77.00 & 72.00 & 64.60 & 55.20 & 82.20 \\
        Skywork-R1V4-30B            & 82.80 & 79.80 & 70.80 & 59.30 & 88.00 \\
        \midrule
        S1-VL-32B-SFT               & 85.00 & 85.10 & 72.00 & 60.80 & 90.00 \\
        S1-VL-32B-RL                & \textbf{91.38} & \textbf{93.50} & \textbf{77.70} & \textbf{67.10} & \textbf{92.70} \\
        \bottomrule
    \end{tabular*}
\end{table}

\subsection{Ablation Studies}
\label{sec:exp_ablation}

To validate the effectiveness of each key design choice in S1-VL, we conduct ablation experiments along two axes: the training strategy and the data construction pipeline. All ablation models are trained under the same hyperparameter settings and evaluated on representative benchmarks spanning both scientific reasoning and Thinking-with-Images tasks. Results are presented in Table~\ref{tab:ablation}.

\subsubsection{Impact of Training Strategy}

We first investigate the contribution of each reinforcement learning stage by progressively removing Stage 3 (Scientific Reasoning RL) and Stage 4 (Thinking-with-Images RL) from the full S1-VL-32B-RL pipeline.

\noindent\textbf{Effect of Stage 3: Scientific Reasoning RL.}
When Stage 3 is skipped and training proceeds directly from the Stage 2 SFT checkpoint into Stage 4 (see row 2), a substantial performance drop is observed on scientific reasoning benchmarks: GMAI drops from 62.13 to 59.63 and Galaxy-10 from 51.52 to 42.45. This directly demonstrates the indispensable role of RL exploration in tackling challenging scientific reasoning tasks that lie beyond the reach of SFT alone. Notably, a moderate decline is also observed on Thinking-with-Images-centric benchmarks (HRBench-4K: 91.38$\to$90.00, MME-RW-CN: 77.70$\to$76.20), indicating that the stronger reasoning foundation cultivated in Stage 3 provides a positive transfer effect to the visual tool-use capabilities developed in Stage 4. The performance gap is particularly pronounced on high-difficulty problems, highlighting that difficult samples require the exploratory optimization unique to RL to break through the SFT performance ceiling.

\noindent\textbf{Effect of Stage 4: Thinking-with-Images RL.}
When Stage 4 is skipped and the Stage 3 checkpoint is evaluated directly (see row 3), scientific reasoning benchmarks remain largely unaffected (GMAI: 62.10, Galaxy-10: 49.55), but Thinking-with-Images-centric benchmarks exhibit a substantial decline (HRBench-4K: 91.38$\to$86.40, MME-RW-CN: 77.70$\to$73.60). While imitation learning in the SFT stage enables the model to acquire the basic form of tool invocation, it cannot guarantee the robustness or effectiveness of tool-use behavior. Stage 4 addresses this limitation by applying reinforcement learning with a carefully designed composite reward function encompassing format, correctness, consistency, efficiency, and tool bonus terms. As discussed in the reward hacking analysis, the formulation of the reward function is critical: a poorly designed reward leads the policy to collapse into degenerate tool-avoidance behavior, whereas a well-calibrated composite reward guides the model to invoke tools at the right moment and in the right manner. The fact that scientific reasoning performance is preserved further confirms that Stage 4 Thinking-with-Images reinforcement learning is highly task-specific—it substantially enhances visual tool-use capability without compromising the scientific reasoning competence established in prior stages.

\subsubsection{Impact of Data Quality}

We next examine the contribution of key data construction components by ablating the quality filtering and adaptive routing mechanisms within the S1-VL-32B-SFT pipeline.

\noindent\textbf{Effect of Thinking-with-Images Data Filtering.}
When the six-dimensional quality filtering pipeline is removed and the model is trained directly on raw, unfiltered Thinking-with-Images trajectory data (see row 5), performance degrades consistently across all benchmarks compared to the filtered baseline (see row 4). Thinking-with-Images-centric benchmarks suffer notably (HRBench-4K: 85.00$\to$80.25, MME-RW-CN: 72.00$\to$69.51), directly corroborating the adverse impact of data noise on visual tool-use competence. Importantly, scientific reasoning benchmarks also deteriorate (GMAI: 58.40$\to$55.17, Galaxy-10: 42.45$\to$40.40), demonstrating that the corrupting effect of noisy supervision signals propagates broadly across reasoning capabilities.

Systematic noise in raw Thinking-with-Images trajectories manifests in several principal forms: \textbf{(i) Redundant tool invocations}—the model repeatedly crops the same or heavily overlapping regions, generating intermediate images that carry no additional visual information; \textbf{(ii) Invalid intermediate images}—certain cropped regions yield blank, severely blurred, or content-absent images, causing subsequent reasoning steps to be grounded in erroneous visual context; \textbf{(iii) Reasoning–action inconsistency}—the natural-language reasoning and the executed tool call are semantically misaligned, transmitting contradictory supervision signals to the policy; and \textbf{(iv) Malformed tool calls}—missing parameters, out-of-bound coordinates, and non-compliant output formats introduce structural noise. Training on such noisy trajectories distorts the model's learned representation of effective tool-use behavior: redundant crops cause the model to internalize repeated invocation as a valid pattern, invalid images encourage shortcut strategies that bypass visual feedback, and reasoning–action inconsistencies inject opposing gradients that induce training instability. The six-dimensional filtering procedure ensures that training trajectories provide clear, consistent, and high-quality supervision, yielding substantial improvements in tool invocation timing, operational precision, and inferential coherence.

\noindent\textbf{Effect of Adaptive Data Routing.}
When the adaptive routing mechanism is removed and all filtered Thinking-with-Images samples are used for training indiscriminately regardless of visual information gain (see row 6), the most severe degradation is observed on scientific reasoning benchmarks (GMAI: 58.40$\to$51.35, Galaxy-10: 42.45$\to$34.73), with Thinking-with-Images-centric benchmarks also declining (HRBench-4K: 85.00$\to$84.00, MME-RW-CN: 72.00$\to$70.82). Without routing, the model exhibits two coupled degradations in tool-use behavior. First, \textbf{inflated invocation frequency}: the model frequently initiates tool calls in scenarios requiring no visual assistance, yet the retrieved visual information contributes negligibly to the final answer. Second, \textbf{degraded invocation quality}: paradoxically, in scenarios that genuinely demand precise visual grounding, the model fails to focus on high-value regions, resulting in decreased cropping precision and less informative intermediate images. The pronounced decline on scientific reasoning benchmarks is particularly noteworthy—it reveals that indiscriminate Thinking-with-Images training causes the model to persistently attempt tool invocations in settings where pure language reasoning is sufficient, interrupting otherwise coherent reasoning chains and introducing unnecessary visual detours. By routing low-gain samples to the pure reasoning training mode, the adaptive routing mechanism teaches the model to judge \emph{when} tool invocation is necessary rather than treating Thinking with Images as a default strategy.

\subsubsection{Overall Analysis}

The ablation results collectively validate the design rationale of S1-VL, demonstrating that each component contributes a distinct and complementary capability.

At the data construction level, quality filtering and adaptive routing are both indispensable: the filtering mechanism eliminates noisy trajectories that would otherwise corrupt tool-use learning, while the routing mechanism ensures the model learns to judge when tool invocation is necessary rather than treating Thinking with Images as a default strategy. At the training level, Stage 3 Scientific Reasoning RL breaks through the performance ceiling imposed by SFT on challenging scientific tasks, while Stage 4 Thinking-with-Images RL, guided by a carefully designed composite reward function, cultivates robust visual tool-use capability that SFT alone cannot reliably produce.

These findings demonstrate that the four-stage training pipeline is not a mere engineering stack, but a principled progressive design: each stage builds targeted capabilities upon the foundation of its predecessor, forming an organic progression of complementary and mutually reinforcing competencies.

\begin{table}[t]
    \centering
    \caption{Ablation study results. Each row removes or modifies one key component of S1-VL-32B. Results are reported on representative benchmarks from both scientific reasoning and Thinking-with-Images directions.}
    \label{tab:ablation}
    \small
    \begin{tabular*}{\textwidth}{cl @{\hspace{1.5em}} l @{\extracolsep{\fill}} cccc}
        \toprule
        \textbf{\#} & \textbf{Configuration} & \textbf{Stage} & \textbf{GMAI} & \textbf{Galaxy-10} & \textbf{HRBench-4K} & \textbf{MME-RW-CN} \\
        \midrule
        1 & S1-VL-32B-RL  & 1,2,3,4 & \textbf{62.13} & \textbf{51.52} & \textbf{91.38} & \textbf{77.70} \\
        2 & ~~w/o Scientific Reasoning RL & 1,2,4 & 59.63 & 42.45 & 90.00 & 76.20 \\
        3 & ~~w/o Thinking-with-Images RL & 1,2,3 & 62.10 & 49.55 & 86.40 & 73.60 \\
        \midrule
        4 & S1-VL-32B-SFT & 1,2 & 58.40 & 42.45 & 85.00 & 72.00 \\
        5 & ~~w/o Data Filtering & 1,2 & 55.17 & 40.40 & 80.25 & 69.51 \\
        6 & ~~w/o Data Adaptive Routing & 1,2 & 51.35 & 34.73 & 84.00 & 70.82 \\
        \bottomrule
    \end{tabular*}
\end{table}


\section{Case Study}
\label{sec:case}

We present representative case studies to provide qualitative insight into the behavior of S1-VL-32B under both reasoning paradigms. Detailed figures with annotated reasoning traces are provided in the appendix.

\noindent\textbf{Thinking-with-Images Success Cases.}
\label{sec:case_success}
We observe that Thinking-with-Images is particularly beneficial when successful problem solving depends on precise local inspection, geometric measurement, or visually grounded computation that cannot be reliably extracted from a single global image encoding. Across diverse scientific domains---including radiology (Figure~\ref{fig:s1-vl-case-twi-biology}), remote sensing (Figure~\ref{fig:s1-vl-case-twi-geography}), and crystallographic diffraction analysis (Figure~\ref{fig:s1-vl-case-twi-chem})---S1-VL autonomously identifies regions of interest, generates targeted crop-and-zoom operations, and integrates the resulting intermediate images into subsequent reasoning steps. These cases demonstrate that the model has learned not merely to invoke tools, but to do so in a spatially precise and task-relevant manner, effectively overcoming the resolution and attention bottlenecks inherent in static full-image encoding.

\noindent\textbf{Scientific Reasoning Cases.}
\label{sec:case_reasoning}
Beyond Thinking-with-Images, S1-VL also demonstrates strong capability in solving complex scientific problems through pure structured reasoning. As illustrated in a multi-image mechanics derivation (Figure~\ref{fig:s1-vl-case-reasoning}), a number-strip algebraic problem (Figure~\ref{fig:s1-vl-case-math-reasoning}), and a galaxy morphology classification task (Figure~\ref{fig:s1-vl-case-galaxy-reasoning}), the model produces rigorous chain-of-thought reasoning---including geometric derivation, symbolic equation solving, and systematic visual feature analysis---entirely within the language modality, without relying on any external tool. These examples confirm that our scientific reasoning training pipeline effectively equips the model with strong multimodal reasoning capabilities, enabling it to handle diverse and challenging scientific tasks through structured deliberation alone.

\noindent\textbf{Thinking-with-Images Failure Cases.}
\label{sec:case_failure}
Analysis of failure cases reveals two representative limitations of the current approach. The first is \emph{imprecise spatial grounding}: as shown in Figure~\ref{fig:s1-vl-badcase-1}, the model sometimes mis-localizes small objects in cluttered high-resolution scenes, requiring a corrective second turn that wastes one full interaction cycle. While the model's capacity for self-correction within the multi-turn loop is encouraging, improving first-attempt localization accuracy remains an important direction. The second, more concerning failure mode is \emph{spurious success}: as shown in Figure~\ref{fig:s1-vl-badcase-2}, the model produces a correct final answer despite cropping an entirely irrelevant region, relying on language priors rather than genuine visual grounding. Such cases are indistinguishable from true successes under outcome-based evaluation and motivate the development of process-level metrics that assess the quality of intermediate visual operations alongside final answer correctness.

\section{Conclusion}
\label{sec:conclusion}

In this paper, we present S1-VL, a multimodal reasoning model tailored to scientific domains that natively supports two complementary reasoning paradigms: Scientific Reasoning and Thinking-with-Images. To address the pervasive issues of redundant, ineffective, and erroneous visual operations in existing Thinking-with-Images training data, we design a six-dimensional quality filtering pipeline together with an adaptive data routing strategy. We further propose a four-stage progressive training pipeline, combined with SAPO-based reinforcement learning, to systematically build both scientific reasoning capability and interactive visual reasoning capability. Systematic evaluation of S1-VL-32B on 13 benchmarks shows that it achieves state-of-the-art performance on all five Thinking-with-Images benchmarks and outperforms larger comparison models on multiple scientific reasoning benchmarks. We release the full model weights of S1-VL-32B to support future research in scientific multimodal reasoning.

Despite these encouraging results, the current model still faces limitations in precise coordinate grounding for image operations, robust code generation, and fine-grained interpretation of specialized scientific imagery. We highlight three directions for future work. First, more targeted training strategies are needed to address the failure modes identified in our case studies and improve reliability in complex visual manipulation scenarios. Second, it is promising to explore training schemes for long-context, multi-image interleaved inputs, enabling cross-figure and long-document scientific reasoning. Third, integrating Thinking-with-Images with broader agent frameworks may advance multimodal systems toward autonomous scientific analysis.

\section*{Author Contributions}

All contributors of this paper are listed below.

\textbf{Contributors:} 
Qingxiao Li,
Lifeng Xu,
QingLi Wang,
Yudong Bai,
Mingwei Ou,
Shu Hu,
Nan Xu.

\bibliography{colm2024_conference}
\bibliographystyle{colm2024_conference}

\appendix
\clearpage
\section{Appendix}
\subsection{Case Study Figures: Thinking-with-Images Success Cases}
\label{appendix:cases}

\begin{figure}[!htbp]
    \centering
    \includegraphics[width=0.95\linewidth]{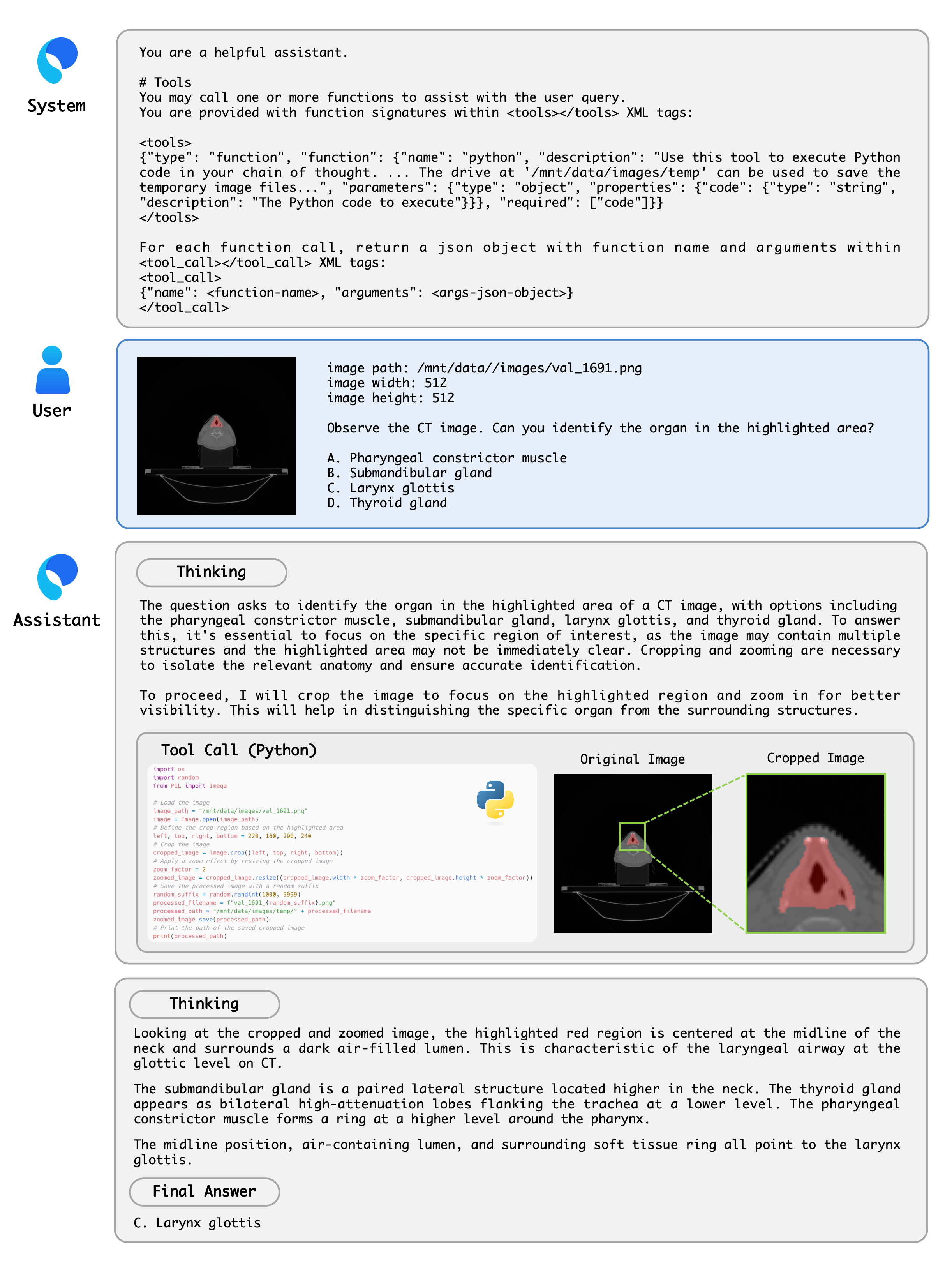}
    \caption{Thinking-with-Images case on a radiology CT image (medical domain). The model crops and zooms into the highlighted anatomical region, isolating a midline structure with a characteristic soft-tissue ring surrounding an air-containing lumen, and correctly identifies it as \textit{Larynx glottis}. This demonstrates how targeted visual manipulation reduces interference from overlapping anatomical structures and enables precise localization.}
    \label{fig:s1-vl-case-twi-biology}
\end{figure}

\begin{figure}[!htbp]
    \centering
    \includegraphics[width=\linewidth]{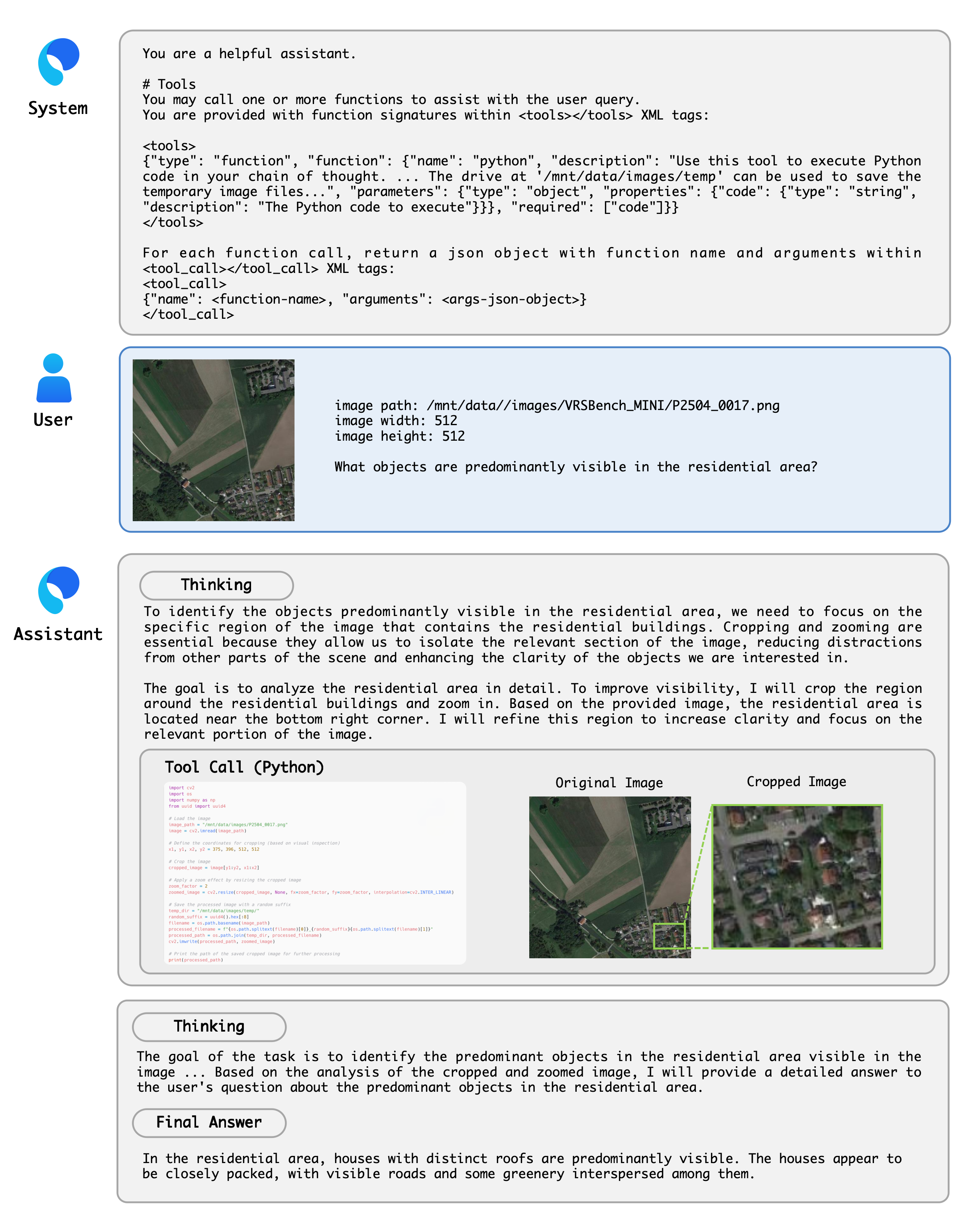}
    \caption{Thinking-with-Images case on a remote sensing image from VRSBench (geography domain). The model localizes the residential cluster in the bottom-right quadrant and generates a crop operation to enlarge it, revealing densely packed houses, road networks, and interspersed greenery that are difficult to resolve at the full-image scale.}
    \label{fig:s1-vl-case-twi-geography}
\end{figure}

\begin{figure}[!htbp]
    \centering
    \includegraphics[width=0.8\linewidth]{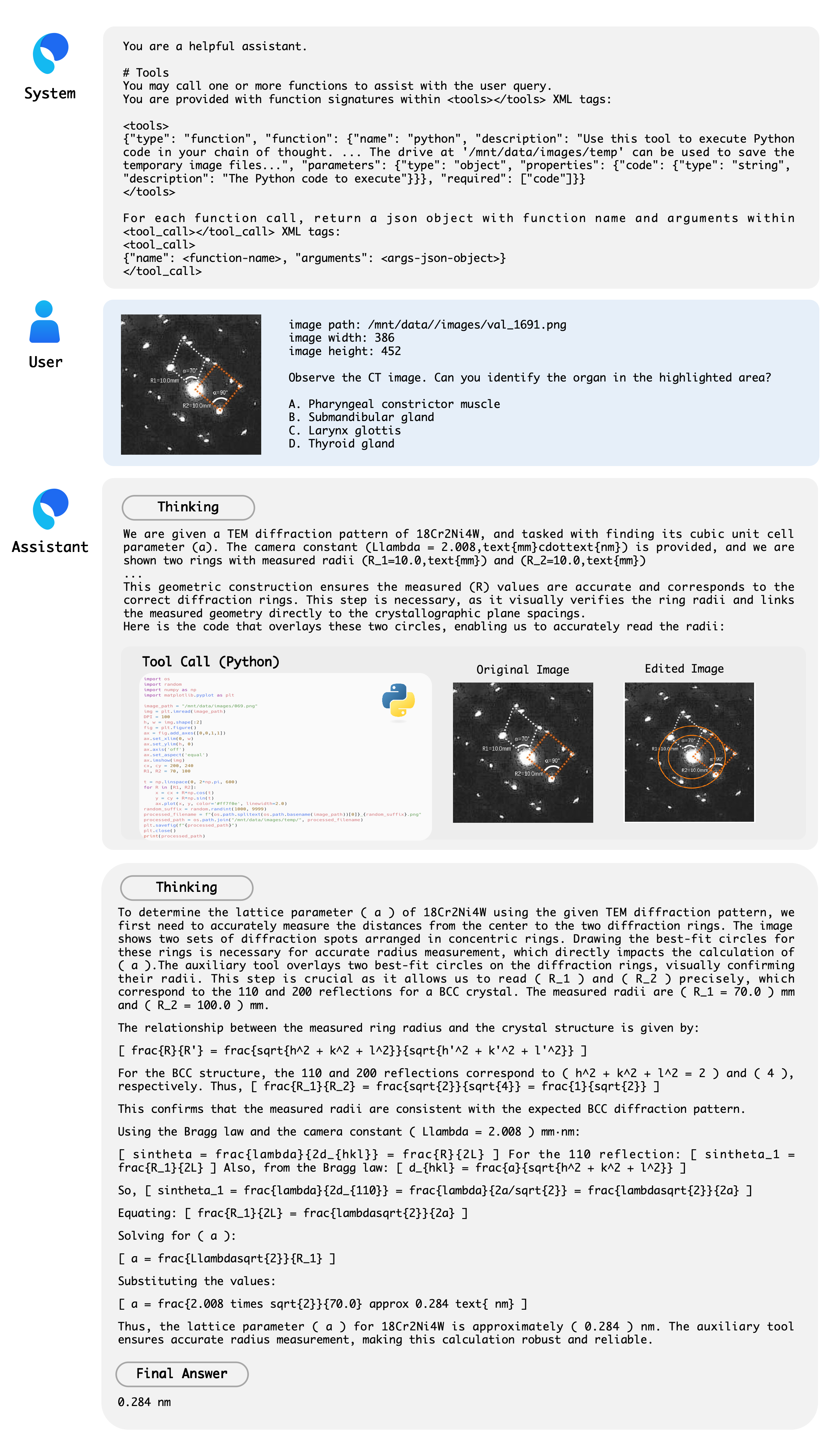}
    \caption{Thinking-with-Images case on TEM diffraction pattern analysis (chemistry/materials science domain). The model overlays best-fit circles onto diffraction rings via a Python tool call, measures ring radii, and applies Bragg's law to derive the BCC lattice parameter $a \approx 0.284\,\text{nm}$ for 18Cr2Ni4W steel, combining tool-augmented visual measurement with crystallographic reasoning.}
    \label{fig:s1-vl-case-twi-chem}
\end{figure}

\clearpage
\subsection{Case Study Figures: Scientific Reasoning Case}

\begin{figure}[!htbp]
    \centering
    \includegraphics[width=1.0\textwidth]{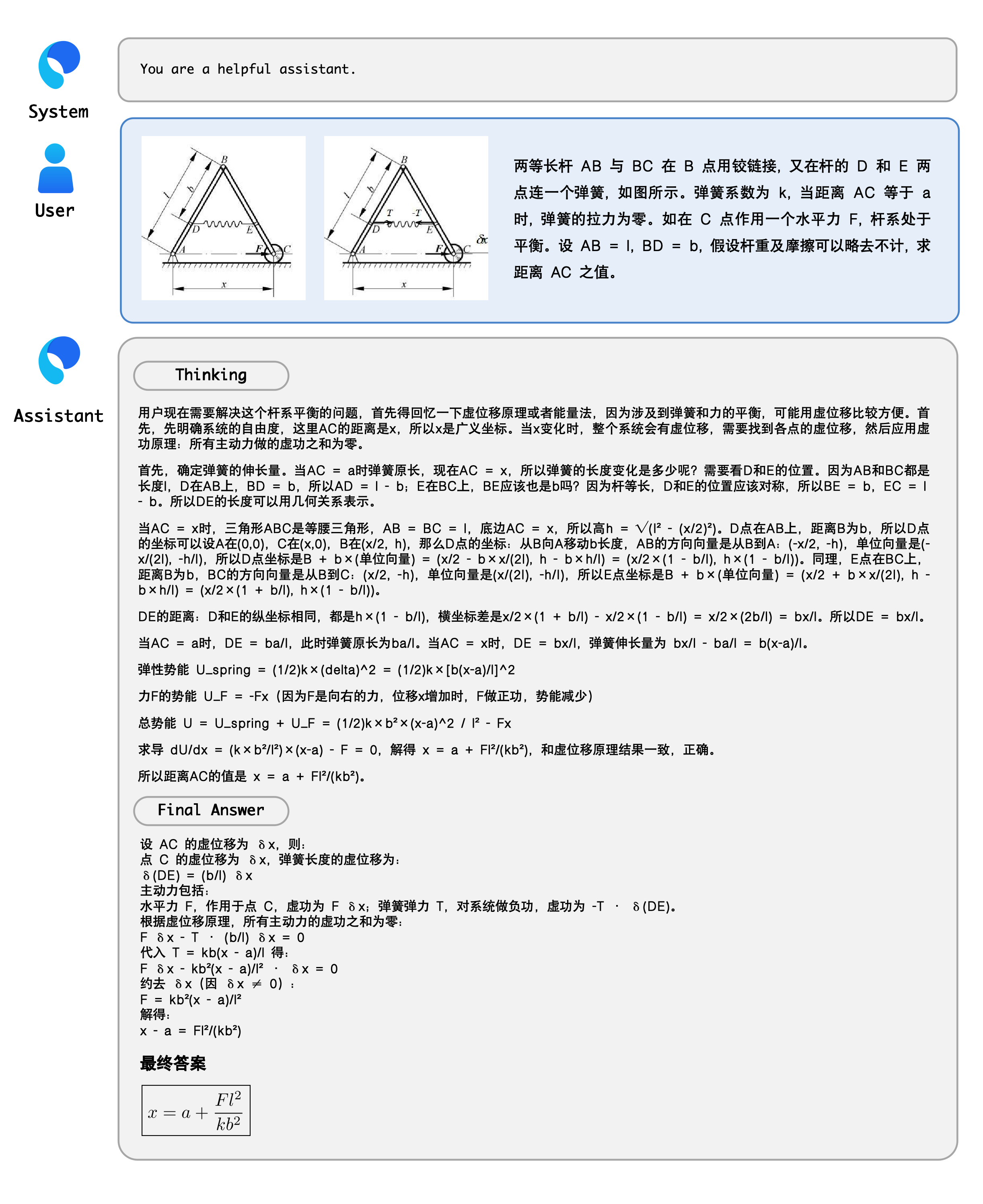}
    \caption{Scientific reasoning case on a multi-image mechanics problem (physics domain). The model jointly analyzes two structural diagrams, derives the equilibrium condition via virtual work and energy methods, and obtains $x = a + \frac{Fl^2}{kb^2}$ through structured chain-of-thought reasoning.}
    \label{fig:s1-vl-case-reasoning}
\end{figure}

\clearpage

\begin{figure}[!htbp]
    \centering
    \includegraphics[width=0.95\textwidth]{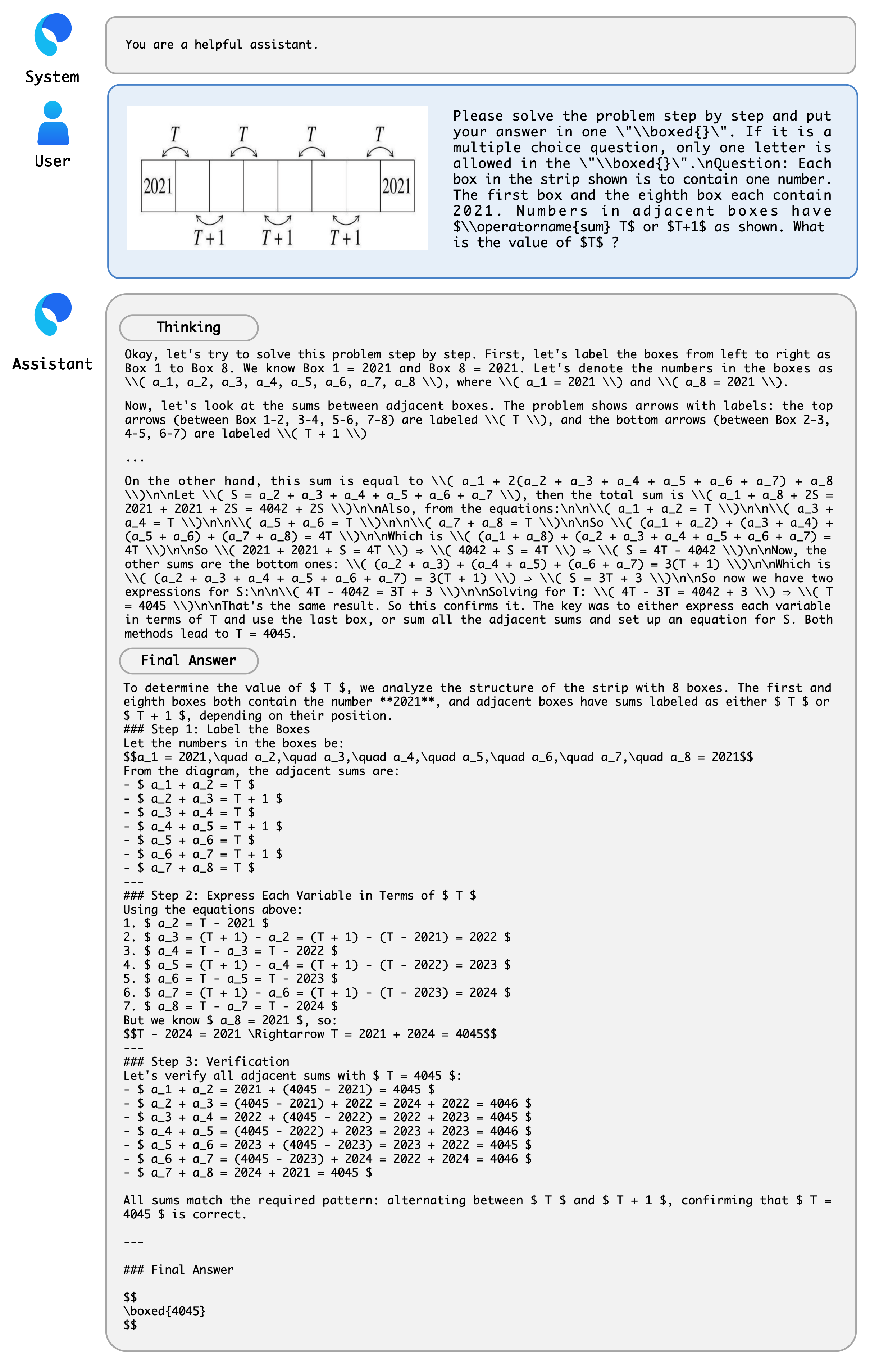}
    \caption{Scientific reasoning case on a number-strip problem (mathematics domain). The model extracts alternating sum constraints from the diagram, expresses box values as functions of $T$, and solves the resulting linear system to obtain $T = 4045$.}
    \label{fig:s1-vl-case-math-reasoning}
\end{figure}

\clearpage

\begin{figure}[!htbp]
    \centering
    \includegraphics[width=1.0\textwidth]{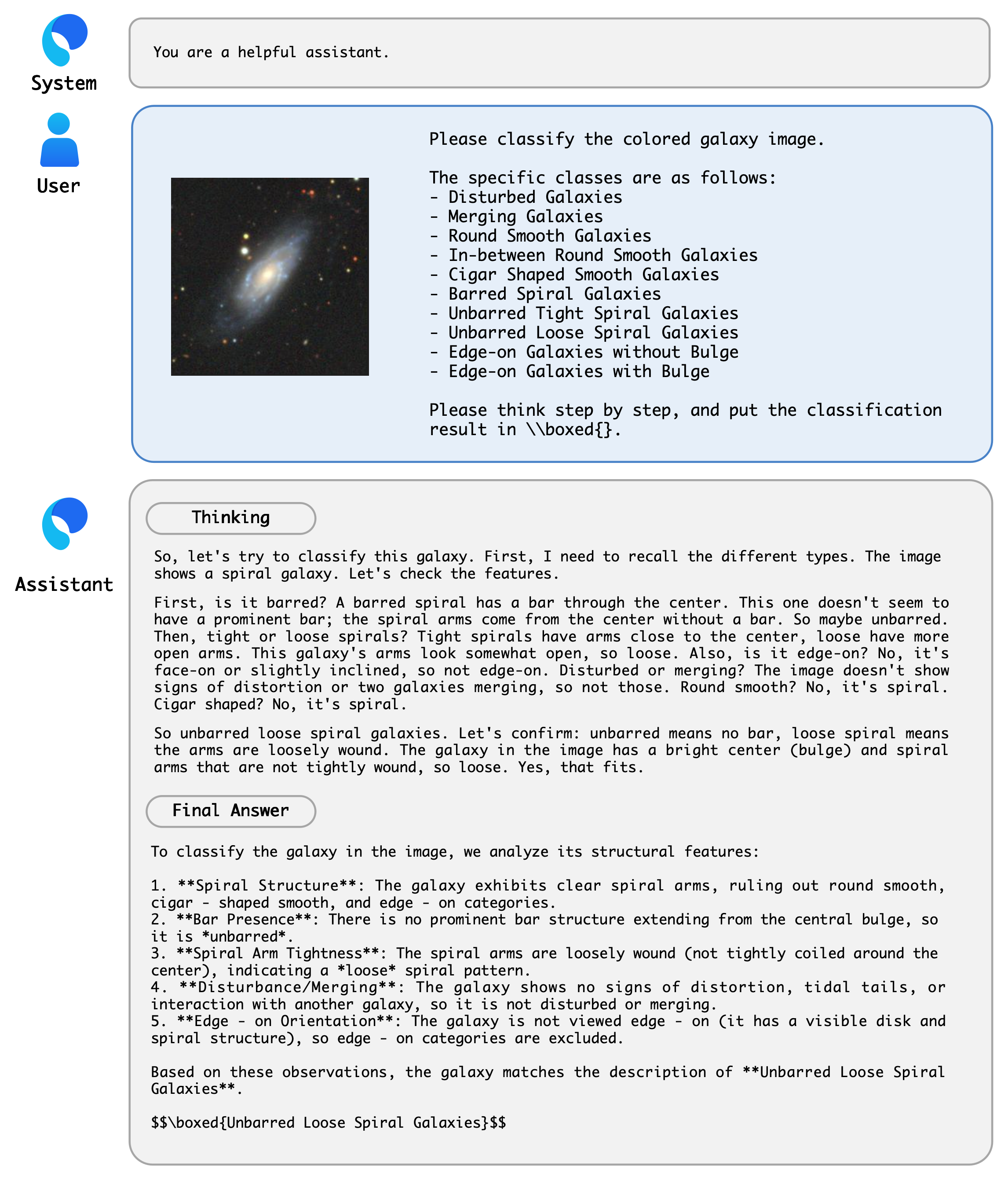}
    \caption{Scientific reasoning case on a galaxy morphology classification task. The model identifies the galaxy as an \textit{Unbarred Loose Spiral Galaxy} via systematic structural analysis of bar presence, spiral arm winding, and disk orientation in Reasoning mode, without invoking any image operation.}
    \label{fig:s1-vl-case-galaxy-reasoning}
\end{figure}

\clearpage
\subsection{Case Study Figures: Thinking-with-Images Failure Cases}

\begin{figure}[!htbp]
    \centering
    \includegraphics[width=0.82\textwidth]{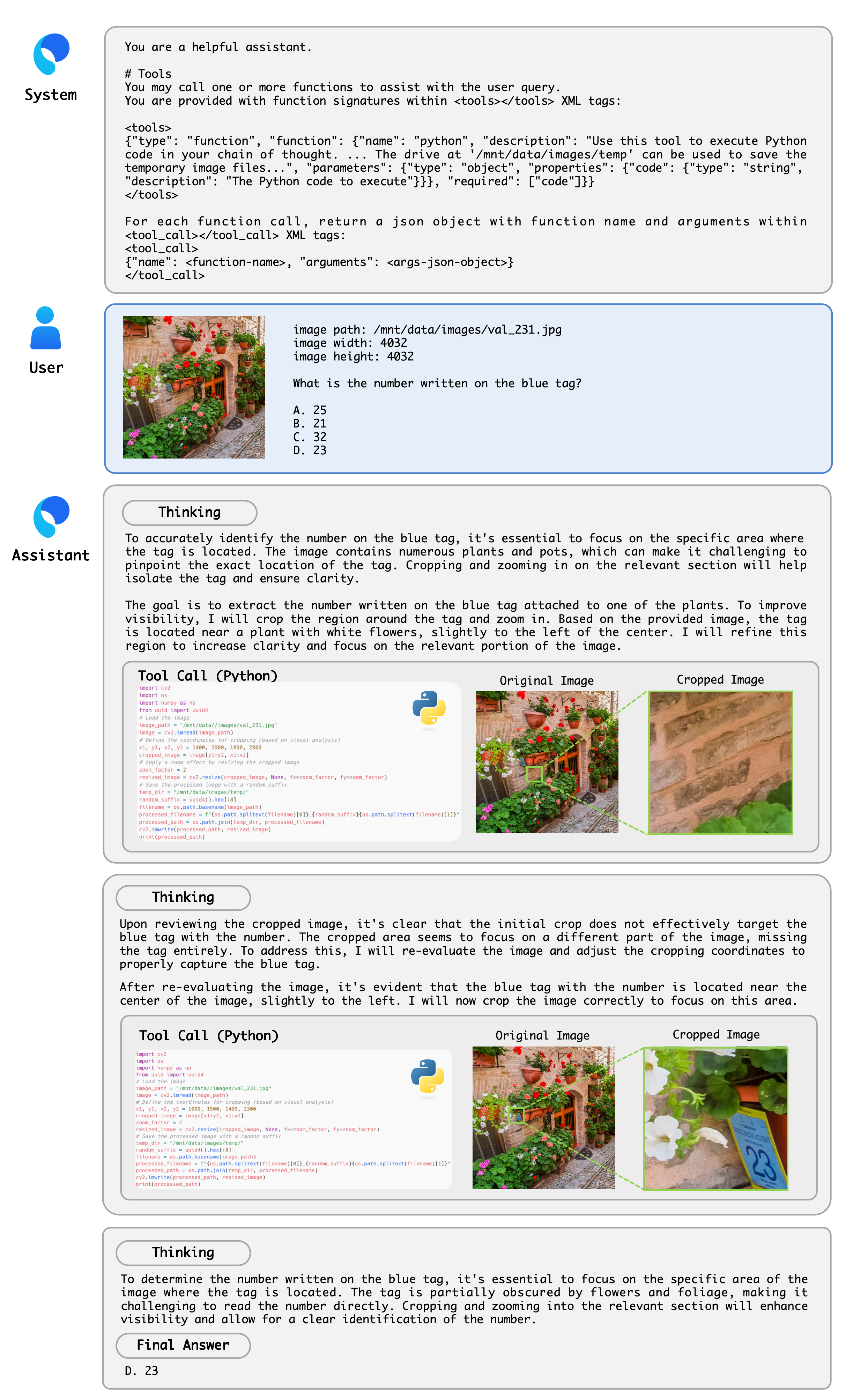}
    \caption{Failure case: imprecise spatial grounding with self-correction. The first crop captures an irrelevant wall region; the model detects the error, re-localizes, and successfully isolates the blue tag on the second attempt, reading the number as 23. This highlights both the self-correction capability and the efficiency cost of erroneous initial localization.}
    \label{fig:s1-vl-badcase-1}
\end{figure}

\begin{figure}[!htbp]
    \centering
    \includegraphics[width=1.0\textwidth]{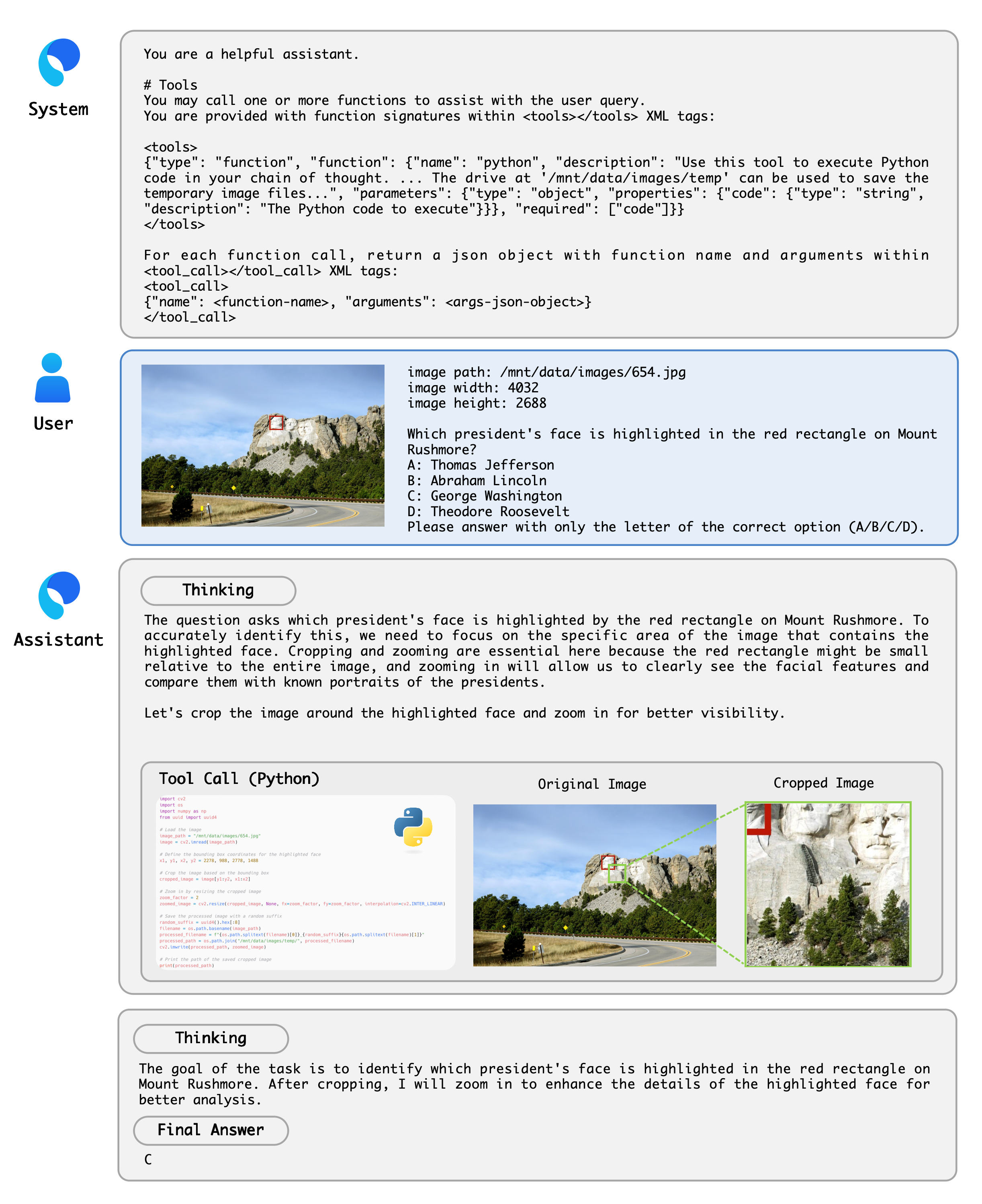}
    \caption{Failure case: spurious success via language priors. The cropped region misses the red-rectangle-marked face entirely, yet the model still outputs the correct answer (George Washington) based on world knowledge rather than visual grounding. This motivates process-level evaluation beyond outcome-based metrics.}
    \label{fig:s1-vl-badcase-2}
\end{figure}

\end{document}